\documentclass[letterpaper]{article} 
\usepackage{aaai2026}  
\usepackage{times}  
\usepackage{helvet}  
\usepackage{courier}  
\usepackage[hyphens]{url}  
\usepackage{graphicx} 
\usepackage{natbib}  
\usepackage{caption} 
\usepackage{algorithm}
\usepackage{algorithmic}

\usepackage{makecell}

\usepackage{newfloat}
\usepackage{listings}
\DeclareCaptionStyle{ruled}{labelfont=normalfont,labelsep=colon,strut=off} 
\floatstyle{ruled}
\newfloat{listing}{tb}{lst}{}
\floatname{listing}{Listing}

\author{
	\textbf{Kewei Chen}\textsuperscript{\rm 1, 2},
	\textbf{Yayu Long}\textsuperscript{\rm 1, 2},
	\textbf{Mingsheng Shang}\textsuperscript{\rm 1, 2}\thanks{Corresponding author.}
}
\affiliations{
	\textsuperscript{\rm 1}Chongqing Institute of Green and Intelligent Technology, Chinese Academy of Sciences\\
	\textsuperscript{\rm 2}Chongqing School, University of Chinese Academy of Sciences\\
	\{chenkewei24, longyayu24\}@mails.ucas.ac.cn, msshang@cigit.ac.cn
}

\usepackage{bibentry}
\usepackage[table]{xcolor}
\usepackage{booktabs}  
\usepackage{multirow}  
\usepackage{amssymb}   
\usepackage{algorithm}
\usepackage{algorithmic}
\usepackage{amsmath}
\usepackage{amssymb}    
\usepackage{algorithm}
\usepackage{algorithmic}
\usepackage{newfloat}
\usepackage{listings}
\DeclareCaptionStyle{ruled}{labelfont=normalfont,labelsep=colon,strut=off} 
\floatstyle{ruled}
\newfloat{listing}{tb}{lst}{}
\floatname{listing}{Listing}
\usepackage{aaai2026}
\usepackage{times}
\usepackage{helvet}
\usepackage{courier}
\usepackage[hyphens]{url}
\usepackage{graphicx}
\usepackage{natbib}
\usepackage{caption}
\title{PIPHEN: Physical Interaction Prediction with Hamiltonian Energy Networks}
\nocopyright
\begin{document}
	
	\maketitle
	
	\begin{abstract}
		Multi-robot systems in complex physical collaborations face a "shared brain dilemma": transmitting high-dimensional multimedia data (e.g., video streams at ~30MB/s) creates severe bandwidth bottlenecks and decision-making latency. To address this, we propose PIPHEN, an innovative distributed physical cognition-control framework. Its core idea is to replace "raw data communication" with "semantic communication" by performing "semantic distillation" at the robot edge, reconstructing high-dimensional perceptual data into compact, structured physical representations. This idea is primarily realized through two key components: (1) a novel Physical Interaction Prediction Network (PIPN), derived from large model knowledge distillation, to generate this representation; and (2) a Hamiltonian Energy Network (HEN) controller, based on energy conservation, to precisely translate this representation into coordinated actions. Experiments show that, compared to baseline methods, PIPHEN can compress the information representation to less than 5\% of the original data volume and reduce collaborative decision-making latency from 315ms to 76ms, while significantly improving task success rates. This work provides a fundamentally efficient paradigm for resolving the "shared brain dilemma" in resource-constrained multi-robot systems.
	\end{abstract}
	
	\maketitle
	\section{Introduction}
	
	Multi-robot systems performing collaborative tasks in complex environments face a "shared brain dilemma": transmitting high-dimensional multimedia perception data (e.g., a 1-second RGB-D video stream can be around 30MB) to a central processing unit causes severe bandwidth bottlenecks and decision-making latency, while fully distributed architectures struggle to maintain global coordination capabilities \cite{dai2024real, dorigo2020reflections, ebrahim2024fully}. This problem is particularly prominent in critical application areas such as industrial automation \cite{nair2024collaborative}, medical surgical assistance \cite{attanasio2021autonomy, zhou2020application}, and agricultural robotics \cite{bonadies2019overview}. We argue that the key to solving this problem lies in shifting from the paradigm of "transmitting raw data" to a new paradigm of "transmitting semantic knowledge."
	
	Existing solutions primarily oscillate between centralized methods that "sacrifice communication for coordination" (e.g., multi-modal fusion \cite{wang2020learning}) and distributed methods that "sacrifice coordination for communication" (e.g., distributed learning \cite{mcmahan2017communication, li2021hermes}). In recent years, although planners based on Large Language Models (LLMs), such as LLaMAR \cite{llamar_paper_2024}, have shown great potential in task decomposition and high-level reasoning, they generally treat robot actions as an "atomic black box," ignoring the physical reality and inter-robot physical coupling behind action execution. They excel at generating logical sequences of "what to do" but fail to answer "how to do it with physical precision," nor do they solve the underlying problem of efficiently sharing the high-dimensional perceptual data required for such precise collaboration. When dealing with complex, highly dynamic physical interaction tasks, these methods generally overlook temporal consistency and uncertainty management \cite{abdar2021review, lakshminarayanan2017simple}, which are critical for achieving robust system performance.
	
	To this end, this paper proposes PIPHEN (Physical Interaction Prediction with Hamiltonian Energy Networks), an innovative distributed physical cognition-control framework designed to address the aforementioned dilemma. Through an elegant perception-cognition-control closed-loop design, it achieves a paradigm shift in information representation. For instance, instead of compressing data files, it distills a 1-second RGB-D raw video stream (approx. 30MB) into structured graph data describing object states and relationships (approx. 1MB), thereby compressing the effective representation of critical information to less than 5\% of the original data volume. First, its core \textbf{Physical Interaction Prediction Network (PIPN)} performs semantic distillation at the robot edge. Subsequently, the Hamiltonian Energy Network (HEN) receives this representation and generates energy-conserving, physically consistent collaborative control commands. The entire framework is efficiently deployed through a three-layer "micro-brain" architecture and utilizes our designed three-stage "Generate-Purify-Deploy" knowledge transformation process to successfully endow resource-constrained edge devices with the physical cognition capabilities of large models.

	The main contributions of this paper include:
	
	(1)\textbf{A novel distributed framework, PIPHEN}: With "semantic distillation" as its core idea, it provides an effective paradigm for solving the "shared brain dilemma" in multi-robot systems;
	
	(2)\textbf{Physical Interaction Prediction Network (PIPN)}: Through hybrid physical representation and physics-constrained modeling, it achieves a precise and compact understanding of physical interactions;
	
	(3)\textbf{Hamiltonian Energy Network (HEN) based on hybrid physical representation}: It uniquely applies the principle of Hamiltonian energy conservation to multi-robot collaborative control based on compact semantic representations, theoretically guaranteeing the physical realism and stability of the control policy;
	
	(4)\textbf{An efficient three-stage LLM knowledge transformation process}: Through "Generate-Purify-Deploy," it successfully deploys the powerful physical cognition capabilities of large models onto resource-constrained robots.
	
	Experimental results show that our method, while drastically reducing communication bandwidth and decision latency (e.g., collaborative decision latency is reduced from 315ms for centralized methods like Concentrative Coordination \cite{yuan2022multi} to 76ms), significantly improves task success rates and control precision.
	
	\begin{figure*}[ht]
		\centering
		\includegraphics[width=1.0\textwidth]{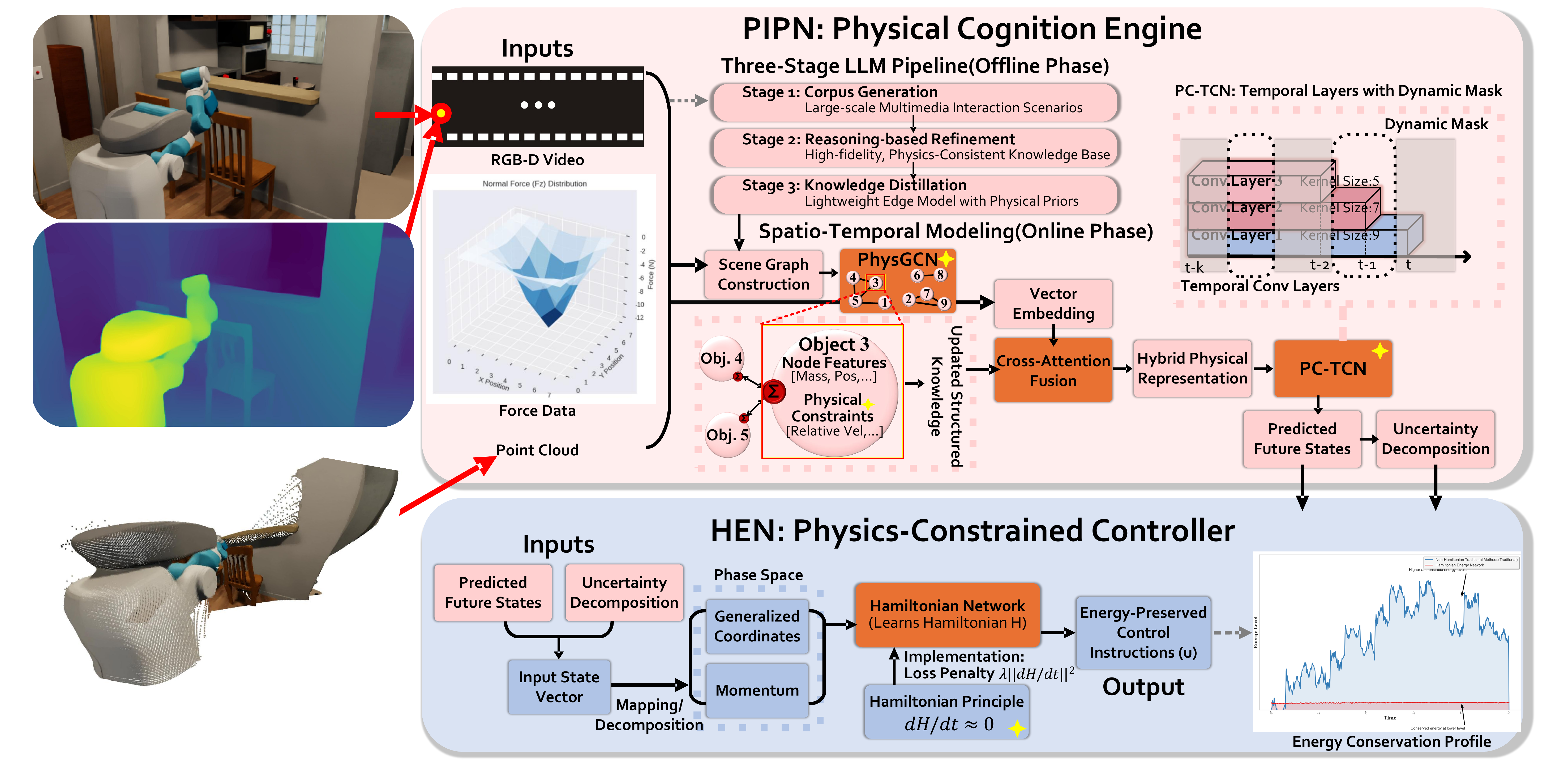}
		\caption{The overall architecture of the PIPHEN framework, comprising two core components: the Physical Cognition Engine (PIPN) and the Physics-Constrained Controller (HEN). The PIPN (top half) is responsible for distilling multi-modal sensory inputs (e.g., RGB-D video, force data, and point clouds) into a compact, structured physical representation, and predicting future physical states and their uncertainties through spatio-temporal modeling. The HEN (bottom half) receives this representation as input and, based on the principle of Hamiltonian energy conservation ($dH/dt \approx 0$), generates physically consistent and stable collaborative control commands. The entire process showcases a complete closed loop from high-dimensional raw perception to low-dimensional semantic knowledge, and then to precise physical control, aimed at efficiently resolving the "shared brain dilemma" in multi-robot systems.}
		\label{fig:piphen_architecture}
	\end{figure*}

	\section{Related Work}
	
	\subsection{Intuitive Physics Understanding}
	An AI's intuitive understanding of the physical world is fundamental to its interaction with the environment. Current research primarily follows three paradigms: structured models understand the world by embedding explicit physical representations in neural networks \cite{xue2023intphys, garrido2025intuitive}, which enhances interpretability but may lack flexibility; pixel-based generative models directly predict future perceptual inputs \cite{gao2022simvp, kipf2019contrastive}, which are highly adaptable but often face challenges in physical consistency and computational efficiency. Recently, representation learning methods have shown great potential. For example, some studies have demonstrated that models can naturally develop an intuitive physical understanding through self-supervised pre-training on natural videos \cite{garrido2025intuitive}. By predicting missing parts of a video in the learned representation space, the model can understand properties like object permanence and shape consistency, proving that models can acquire physical intuition similar to human infants without hard-coded physical knowledge.
	
	\subsection{Physics-based Robot Control}
	Translating physical understanding into effective robot control strategies is an ongoing challenge. In recent years, energy-based control frameworks have shown great potential due to their ability to express system dynamics through conservation principles \cite{greydanus2019hamiltonian, cranmer2020lagrangian}. For example, Hamiltonian Neural Networks (HNNs) can learn system dynamics while adhering to energy conservation. Researchers have further proposed port-Hamiltonian neural ODE networks based on Lie groups \cite{duong2024port}, which embed physical constraints directly into the network architecture, providing a unified method for stabilization and trajectory tracking for various robot platforms. 
	Meanwhile, the application of Physics-Informed Neural Networks (PINNs) has been extended to robot modeling and control, achieving precise control performance by combining traditional and emerging technologies, and has been validated in real-world experiments \cite{liu2024physics}. Additionally, physics-informed neural controllers have been successfully applied to complex tasks such as robotic deployment of deformable linear objects \cite{tong2023sim2real}.
	Unlike other works that apply Hamiltonian principles to multi-robot systems focusing on stabilization or decentralized learning \cite{sebastian2025physics,sebastian2023lemurs,furieri2022distributed}, our work uniquely utilizes the HEN as a collaborative controller that translates high-level, distilled semantic knowledge from the PIPN into energy-conserving actions for complex, physically-coupled tasks, rather than focusing purely on state-based stabilization.

	\subsection{Language Models for Multi-agent Planning}
	Recently, leveraging Large Language Models (LLMs) for multi-agent planning has become a significant research direction. Cutting-edge methods such as RoCo \cite{roco}, CoELA \cite{coela}, as well as SmartLLM \cite{smartllm} and LLaMAR \cite{llamar_paper_2024}, have achieved remarkable success in decomposing complex natural language instructions into logical subtask sequences. They utilize the powerful common-sense reasoning and planning capabilities of LLMs to coordinate the macro-level behaviors of agents. However, these works primarily focus on high-level "task planning" and generally treat low-level "action execution" as a solved, deterministic "black box." As stated in LLaMAR's introduction, its core contribution lies in the planning and reasoning loop, which excels at determining "what to do next." However, for problems involving underlying physical dynamics and multi-body coupling, it lacks effective modeling methods for "how to do it," nor does it address the underlying information-sharing problem necessary to realize the plan.

	\section{Method}
	
	To solve the "shared brain dilemma" in multi-robot collaboration, we propose the PIPHEN distributed physical cognition-control framework. As mentioned in the introduction, the core idea of this framework is to perform "semantic distillation" at the robot edge, transforming high-dimensional perceptual data into low-dimensional, structured physical semantic information, and sharing only this highly compressed knowledge across the network to achieve efficient collaboration. The size of this representation scales linearly with the number of objects in the scene, remaining low in typical industrial assembly scenarios (<50 objects) and ensuring the method's scalability.
	
	\subsection{System Architecture Overview}
	
	PIPHEN adopts a hierarchical "micro-brain" architecture, which, while ensuring global coordination, significantly reduces the collaborative decision latency from 315ms in centralized methods like Concentrative Coordination to just 76ms. This architecture primarily consists of three cooperative layers: the top-level Central Coordination Layer ("Brain") is responsible for global knowledge fusion and generating a collaborative strategy; the middle Local Execution Layer ("Cerebellum") is deployed on each robot, running a lightweight PIPN for real-time perception and a HEN for real-time control; additionally, there is a Specialized Processing Layer ("Micro-Brain"), which provides dynamically loadable function modules that can be invoked across robots, endowing the system with high flexibility and scalability.

	\subsection{Distributed Physical Interaction Prediction Network (PIPN)}
	The PIPN is the perceptual and cognitive core of the system, responsible for efficiently constructing an interpretable and compact model of the physical world from multi-modal data.
	
	\subsubsection{Hybrid Physical Representation}
	To balance interpretability and computational efficiency, PIPN constructs an innovative \textbf{hybrid physical representation}: it fuses a structured \textbf{Physical Knowledge Graph} (representing object properties and relationships) with a \textbf{Task Vector Embedding} generated by a Transformer encoder (capturing dynamics and contextual information). Specifically, these two representations are fused through a Cross-Attention module, where the vector embedding acts as the Query to dynamically aggregate the most relevant physical attributes from the knowledge graph.
	
	\subsubsection{Physical Relationship and Temporal Modeling}
	To accurately model dynamic interactions, we designed a Physics-aware Graph Convolutional Network (PhysGCN). Our PhysGCN consists of L graph convolutional layers, each integrating our proposed relational attention module, which encodes physical constraints (such as mass ratio, relative velocity, etc.) as part of the attention weights, enabling the network to prioritize aggregating information from physically more relevant nodes. Its top-level formula is:
	
	\begin{align}
		R = \text{PhysGCN}(\{f_p^i\}, A, E; \theta_g)
	\end{align}
	where $R$ is the relational representation, $\{f_p^i\}$ are the initial physical features, $A$ is the adjacency matrix, $E$ are the edge features, and $\theta_g$ are the network parameters. Subsequently, a Physics-Consistent Temporal Convolutional Network (PC-TCN) is responsible for predicting the dynamic evolution.

	This network, through our proposed dynamic causal masking mechanism, adaptively adjusts its temporal receptive field to more accurately capture the causal relationships of physical interactions.

	\subsubsection{Loss Function with Energy-Momentum Conservation}
	To guide PIPN in learning physically realistic dynamic predictions, we introduce a regularization term based on physical conservation laws in addition to the traditional prediction loss $\mathcal{L}_{\text{pred}}$. The final loss function $\mathcal{L}$ is a weighted sum:
	\begin{align}
		\mathcal{L} = \mathcal{L}_{\text{pred}} + \lambda_{\text{phy}} \mathcal{L}_{\text{phy}}
	\end{align}
	where $\lambda_{\text{phy}}$ is a hyperparameter set to 0.1.
	
	The prediction loss $\mathcal{L}_{\text{pred}}$ measures the L2 discrepancy between the predicted states (position $p_i$, pose $q_i$) and the true states for $N$ objects over $T$ timesteps:
	\begin{align}
		\mathcal{L}_{\text{pred}} = \frac{1}{N \cdot T} \sum_{t=1}^{T} \sum_{i=1}^{N} \left( || \hat{p}_i^t - p_i^t ||_2^2 + || \hat{q}_i^t - q_i^t ||_2^2 \right)
	\end{align}
	where $\hat{p}_i^t, \hat{q}_i^t$ are the predicted values, and $p_i^t, q_i^t$ are the ground truth values.
	
	The physics-consistency loss $\mathcal{L}_{\text{phy}}$ penalizes violations of energy and momentum conservation:
	\begin{align}
		\mathcal{L}_{\text{phy}} = w_E \mathcal{L}_{E} + w_M \mathcal{L}_{M}
	\end{align}
	\paragraph{Energy Conservation Loss $\mathcal{L}_{E}$:} We approximate the total system energy $E_{\text{total}}^t$ (kinetic and potential)  and penalize its change over time:
	\begin{align}
		E_{\text{total}}^t &= \sum_{i=1}^N \left( \frac{1}{2}m_i (v_i^t)^2 + m_i g h_i^t \right) \\
		\mathcal{L}_{E} &= \frac{1}{T-1} \sum_{t=1}^{T-1} (E_{\text{total}}^{t+1} - E_{\text{total}}^t)^2
	\end{align}
	where $m_i, v_i^t, h_i^t$ are the mass, velocity, and height of object $i$ at time $t$.
	
	\paragraph{Momentum Conservation Loss $\mathcal{L}_{M}$:} For any colliding object pair $(i, j)$, we enforce momentum conservation between the pre-collision ($t_{\text{pre}}$) and post-collision ($t_{\text{post}}$) states:
	\begin{align}
		\mathcal{L}_{M} = \sum_{(i,j) \in \text{collisions}} || (m_i \hat{v}_i^{t_{\text{post}}} + m_j \hat{v}_j^{t_{\text{post}}}) - (m_i v_i^{t_{\text{pre}}} + m_j v_j^{t_{\text{pre}}}) ||_2^2
	\end{align}
	where $\hat{v}$ is the predicted velocity and $v$ is the input velocity . By minimizing $\mathcal{L}$, PIPN learns dynamics that conform to physical laws. (Further details are in Appendix A).
	
	\subsubsection{Large Model-Enhanced Edge Physics Cognition}
	To achieve complex physical reasoning on resource-constrained robots, we designed a "\textbf{Generate-Purify-Deploy}" three-stage knowledge transformation process:
	\begin{enumerate}
		\item \textbf{Large-Scale Knowledge Corpus Generation}: Utilize large generative models (e.g., Claude-3.7-Sonnet) to generate large-scale, diverse interaction scenarios in simulation to address the problem of sparse physical interaction data.
		\item \textbf{Physics-Reasoning-Based Knowledge Purification}: Use a foundation model with strong logical reasoning capabilities (e.g., GPT-4o) to act as a "\textbf{Physics Verifier}," evaluating and filtering the physical consistency of the generated data to build a high-quality "expert knowledge base."
		\item \textbf{Efficient Knowledge Distillation for the Edge}: Finally, distill the purified expert knowledge and "inject" it into a lightweight edge multi-modal model (Qwen2.5-VL-3B) using knowledge distillation techniques.
	\end{enumerate}
	
	\subsubsection{Uncertainty Decomposition and Collaborative Learning}
	To enhance the system's robustness in the real world, we decompose the prediction uncertainty into three parts: perception, model, and environment. 
	We use their linear sum ($U_{\text{total}} = U_{\text{perc}} + U_{\text{model}} + U_{\text{env}}$) as an approximate estimate of the total uncertainty, which is a common simplification in ensemble learning and Bayesian approximation to ensure model tractability \cite{lakshminarayanan2017simple, abdar2021review}. 
	We quantify $U_{\text{perc}}$, $U_{\text{model}}$, and $U_{\text{env}}$ using established methods of Monte Carlo Dropout, Deep Ensembles, and direct distributional prediction, respectively.
	We validate the effectiveness of this design in our ablation studies.
	Concurrently, we employ a hybrid paradigm of federated learning and knowledge distillation for collaborative training.
	\textbf{(The specific methods for quantifying uncertainty and the details of the collaborative learning framework are elaborated in Appendix B).}

	\subsection{Hamiltonian Energy Network (HEN)}
	The Hamiltonian Energy Network (HEN) works in close collaboration with the PIPN and serves as the control execution core of the PIPHEN system. The HEN receives the physical knowledge representation from the PIPN as input, translating physical understanding into precisely coordinated action control while ensuring energy conservation and action stability, thus forming a complete perception-control loop. Its core is based on \textbf{Hamiltonian mechanics}, modeling the state of the entire system and its total energy $H$, and ensuring the physical consistency of control in the following form:
	\begin{align}
		\dot{x} = f(x, u) \quad \text{s.t.} \quad \frac{dH}{dt} \approx 0
	\end{align}
	where $x$ is the system's state vector, which includes generalized coordinates and momentum. In Hamiltonian mechanics, these two sets of variables together define the system's Phase Space, a "map" that can completely describe all dynamic states of the system. The core task of HEN is to learn the system dynamics that follow energy conservation within this phase space. And $u$ is the control command vector. The constraint $\frac{dH}{dt} \approx 0$ requires that the control commands must conserve the total energy of the system, thereby fundamentally guaranteeing the smoothness, stability, and physical realism of the coordinated actions. 
	We train the HEN policy using Imitation Learning (Behavior Cloning). Expert data is generated in a two-stage process: first, a PDDL planner creates symbolic actions, which TrajOpt then converts into optimal physical trajectories.
	
	In implementation, we guide the network to learn an energy-conserving control policy by adding a penalty term $\lambda ||\frac{dH}{dt}||^2$ to the Imitation Learning loss function, thus achieving efficient Energy-based coordination between the central layer and local robots. 
	(The specific network structure of HEN, detailed training process, and implementation details of the Hamiltonian equations are elaborated in Appendix C).

	In summary, in a typical workflow, the PIPN first distills a hybrid representation containing physical knowledge from multi-modal data and predicts its dynamic evolution. Subsequently, the HEN receives this compact representation and generates energy-conserving, physically consistent collaborative control commands. Finally, these commands are distributed to each robot for execution through the underlying distributed communication mechanism. This perception-cognition-control closed-loop design ensures the efficiency, robustness, and physical realism of the entire system in complex physical interactions.
	
	\subsection{Distributed Communication and Sharing}
	
	PIPHEN's efficient operation relies on its innovative communication mechanism. We adopt a \textbf{hierarchical communication strategy}, dividing physical knowledge into a "core semantic layer" that must be shared in real-time and a "detailed supplementary layer" that is transmitted on demand. In addition, the system introduces an \textbf{incremental update mechanism}, transmitting only the changes in physical knowledge to avoid redundant information transfer. To achieve efficient selective sharing, we also designed an information value assessment module, which decides which knowledge is most worthy of transmission based on its task relevance, novelty, and redundancy. \textbf{(The specific implementation and quantification methods of the information value assessment module are detailed in Appendix D).}
	
	At the sharing mechanism level, each robot maintains a local physical knowledge base and uses technologies like Distributed Hash Tables (DHT) to achieve efficient cross-robot knowledge retrieval and indexing. This mechanism not only promotes physical cognition sharing among robots but also further enhances information access efficiency and system fault tolerance.
	
	In summary, in a typical workflow, the PIPN first distills a hybrid representation containing physical knowledge from multi-modal data and predicts its dynamic evolution. Subsequently, the HEN receives this compact representation and generates energy-conserving, physically consistent collaborative control commands. Finally, these commands are distributed to each robot for execution through the underlying distributed communication mechanism. This perception-cognition-control closed-loop design ensures the efficiency, robustness, and physical realism of the entire system in complex physical interactions.
	
	\section{Experiments}
	
	To comprehensively evaluate the performance of the PIPHEN framework and directly compare it with state-of-the-art multi-agent methods based on Large Language Models (LLMs), we adopted the highly challenging experimental setup used by LLaMAR \cite{llamar_paper_2024}. Our core objective is to demonstrate that by endogenizing physical cognition and energy conservation principles into the decision-making process, PIPHEN can surpass SOTA methods that primarily rely on LLMs for posterior reasoning on these complex benchmarks.
	
	\subsection{Experimental Setup}
	
	\subsubsection{MAP-THOR Benchmark}
	To evaluate the performance of our method and benchmark it against other baselines, we adopted the MAP-THOR (Multi-Agent Planning tasks in AI2-THOR) benchmark dataset. 
	Although Smart-LLM \cite{smartllm} introduced a dataset of 36 tasks categorized by complexity in AI2-Thor \cite{ai2thor}, these tasks are limited to single-floor layouts. 
	This limitation hinders the testing of a planner's robustness across different room layouts. 
	
	In contrast, MAP-THOR includes tasks that can be solved by single or multiple agents. 
	We categorize tasks into four classes based on the ambiguity of the language instructions. 
	To test the planner's robustness, we provide five different floor plans for each task. 
	We also integrated an auto-check module to verify sub-task completion and evaluate planning quality. 
	This dataset contains 45 tasks, each defined for five unique floor plans, ensuring comprehensive testing and evaluation. 
	For this task, the PIPN is specifically trained to predict the future 6D pose (position and orientation) and velocity for all relevant object nodes in the scene.
	
	We conducted experiments on tasks of varying difficulty levels, where an increase in task difficulty corresponds to an increase in the ambiguity of language instructions: from explicitly specifying item type, quantity, and target location (e.g., \texttt{put the bread, lettuce, and one tomato in the fridge}), to progressively omitting item quantity (e.g., \texttt{put all apples in the fridge}), omitting item type and quantity (e.g., \texttt{put all groceries in the fridge}), and finally, omitting all elements entirely (e.g., \texttt{clean the floor}). 
	A complete list of categorized tasks can be found in Appendix E.

	\subsubsection{Search and Rescue (SAR) Environment}
	To demonstrate the effectiveness of PIPHEN over explicit coordination in multi-agent settings, we evaluated it in a partially observable grid-world search and rescue and fire-fighting environment. 
	Depending on the scenario, the environment contains a mix of missing persons to be found and wildfires that must be extinguished before they spread. 
	In this grid-world setting, the PIPN is modeled to predict the future values of grid cells, such as the spread of fire or changes in cell state.
	More details on this environment are provided in Appendix F.
	
	All simulation experiments were conducted in the AI2-THOR environment, supported by our 4-card A800-80G server (see Appendix G for details). Meanwhile, more physics-fidelity-focused experiments conducted in NVIDIA Isaac Sim are provided as supplementary material.

	\subsection{Evaluation Metrics and Baseline Methods}
	
	\subsubsection{Evaluation Metrics}
	We use the following widely recognized metrics in the multi-agent planning domain to comprehensively evaluate algorithm performance: \textbf{Success Rate (SR, \%)}, the proportion of trials where all sub-tasks were successfully completed; \textbf{Transport Rate (TR, \%)}, the proportion of sub-tasks completed in a single task trial, providing a finer granularity of task completion; \textbf{Coverage (C, \%)}, the proportion of successful interactions with target objects, a metric particularly useful in scenarios where objects are implicitly specified; \textbf{Balance (B)}, which measures the balance of successful actions performed by each agent, defined as $\min\{s_i\}/\max\{s_i\}$, where 1 indicates perfect balance; and \textbf{Average Steps (S)}, the number of high-level actions taken by the team to complete the task, capped at 30. For all evaluation metrics, we report the average values across tasks. To ensure the rigor of our results, detailed results for all core experiments, including 95\% confidence intervals, are provided in Appendix H. For binomial distribution metrics like Success Rate (SR), we use the Clopper-Pearson method to calculate the confidence intervals.
	
	\subsubsection{Baseline Methods}
	We compare PIPHEN against the full suite of SOTA baseline methods used in the LLaMAR paper, including LLaMAR itself. These baselines include: \textbf{Act}/\textbf{ReAct}/\textbf{CoT}, representing different prompt engineering strategies for LLM Agents; \textbf{SmartLLM} \cite{smartllm}, an LLM Agent employing a "plan-and-execute" paradigm; \textbf{CoELA} \cite{coela}, a decentralized multi-agent LLM framework; and \textbf{LLaMAR} \cite{llamar_paper_2024}, one of the current state-of-the-art modular cognitive architectures that integrates planning, execution, and correction through specialized LLM roles. The implementation details of all baseline methods follow their original papers to ensure a fair comparison.
	
	\subsection{Experimental Results and Analysis}
	
	\subsubsection{Analysis of Model Selection in the Knowledge Transformation Process}
	To investigate the dependency and robustness of PIPHEN's core "Generate-Purify-Deploy" three-stage knowledge transformation process on the underlying large models, we adopted the experimental approach from LLaMAR and tested the framework's performance using different models at each stage. As shown in Table \ref{tab:model_selection_ablation}, we compared the performance of different model combinations.
	
	\begin{table}[h!]
		\centering
		\caption{The impact of different model selections in the "Generate-Purify-Deploy" process on PIPHEN's performance.}
		\label{tab:model_selection_ablation}
		\resizebox{\linewidth}{!}{
			\begin{tabular}{lcccc}
				\toprule
				\textbf{Model Config (Generate/Purify/Deploy)} & \textbf{SR (\%)} $\uparrow$ & \textbf{TR (\%)} $\uparrow$ & \textbf{C (\%)} $\uparrow$ & \textbf{B} $\uparrow$ \\
				\midrule
				\textbf{Default (Claude-3.7 / GPT-4o / Qwen2.5-VL)} & \textbf{75} & \textbf{95} & \textbf{98} & \textbf{0.89} \\
				GPT-4o / GPT-4o / Qwen2.5-VL & 73 & 93 & 97 & 0.88\\ 
				Claude-3.7 / Claude-3.7 / Qwen2.5-VL & 70 & 90 & 96 & 0.85 \\
				Claude-3.7 / GPT-4o / Qwen2.5-0.5B & 68 & 88 & 94 & 0.86\\
				\bottomrule
			\end{tabular}
		}
	\end{table}
	
	The experimental results clearly show that although replacing models leads to some performance degradation, the overall framework of PIPHEN exhibits good robustness. Among the stages, "Purify" and "Deploy" are more sensitive to model capability, which aligns with our design expectations: the "Purify" stage requires strong logical reasoning ability to ensure the physical reality of the knowledge, while the edge model in the "Deploy" stage directly determines the system's perceptual and cognitive upper limit at the terminal. Even with sub-optimal model combinations, PIPHEN's performance is still significantly better than most baseline methods, which proves the effectiveness of its framework design, rather than merely relying on a specific powerful model.

	\subsubsection{Comparison with State-of-the-Art Methods}
	Table \ref{table:baselines_comparison} clearly shows the performance comparison of PIPHEN with all baseline methods, including LLaMAR. The results are significant: PIPHEN achieves the best performance on all key metrics.
	
	\begin{table}[h!]
		\centering
		\caption{Performance comparison of PIPHEN with SOTA baseline methods in the 2-agent MAP-THOR scenario. For all metrics, higher is better (except for Steps, where lower is better). The best results are shown in bold. The values in parentheses are the 95\% confidence intervals.}
		\resizebox{0.98\linewidth}{!}{
			{\footnotesize
				\vspace{-2mm}
				\begin{tabular}{cccccc} 
					\toprule
					\textbf{Methods} & \textbf{Success}   & \textbf{Transport}& \textbf{Coverage} & \textbf{Balance} & \textbf{Steps}\\
					& \textbf{Rate (\%)} & \textbf{Rate (\%)} & \textbf{(\%)} & & \\
					\midrule
					\multirow{2}{*}{Act} & 33 & 67   & 91 & 0.59 & 24.8  \\ 
					& \tiny{(19, 49)} & \tiny{(59, 76)} &  \tiny{(86,95)} &  \tiny{(0.52, 0.66)} &  \tiny{(22.1,27.7)}  \\ 
					\multirow{2}{*}{ReAct} & 34 & 72 & 92 & 0.67 & 24.3\\
					& \tiny{(20, 50)} & \tiny{(64, 81)} & \tiny{(87, 96)} & \tiny{(0.59, 0.74)} & \tiny{(21.5, 27.2)}\\
					\multirow{2}{*}{CoT} & 14 & 59 & 87 & 0.62 & 26.9\\
					& \tiny{(6, 25)} & \tiny{(48, 71)} & \tiny{(80, 93)} & \tiny{(0.54, 0.71)} & \tiny{(24.1, 29.8)}\\
					\multirow{2}{*}{SmartLLM} & 11 & 23 & 91 & 0.45 & 28.5\\
					& \tiny{(4, 21)} & \tiny{(14, 35)} & \tiny{(85, 96)} & \tiny{(0.37, 0.54)} & \tiny{(25.8, 30.0)}\\
					\multirow{2}{*}{CoELA} & 25 & 46 & 76 & 0.73 & 25.7\\
					& \tiny{(13, 40)} & \tiny{(35, 58)} & \tiny{(68, 84)} & \tiny{(0.65, 0.82)} & \tiny{(22.9, 28.6)}\\
					\multirow{2}{*}{LLaMAR} & 66 & 91 & 97 & 0.82 & 21.9\\
					& \tiny{(50, 80)} & \tiny{(83, 96)} & \tiny{(93, 99)} & \tiny{(0.75, 0.89)} & \tiny{(18.8, 26.4)}\\
					\midrule
					\multirow{2}{*}{\textbf{PIPHEN (ours)}} & \textbf{75} & \textbf{95} & \textbf{98} & \textbf{0.89} & \textbf{20.1} \\ 
					& \tiny{\textbf{(61, 86)}} & \tiny{\textbf{(89, 99)}} &  \tiny{\textbf{(94,100)}} &  \tiny{\textbf{(0.82, 0.95)}} &  \tiny{\textbf{(17.3, 23.2)}}  \\ 
					\bottomrule
				\end{tabular}
		}}
		
		\label{table:baselines_comparison}
	\end{table}

	Analyzing the fundamental reason, although LLaMAR achieves powerful planning capabilities through its modular LLM roles, it is essentially "reactive"—it relies on posterior reasoning from current visual-language information to decide the next action, lacking "foresight" into the evolution of physical dynamics. When a task requires precise prediction of the consequences of physical interaction (e.g., where an object will roll to after being pushed), LLaMAR's LLM module can only provide a vague judgment based on common sense.
	
	In contrast, our PIPHEN, with its Physical Interaction Prediction Network (PIPN), can construct a precise, differentiable model of the physical world, thus \textbf{predicting} rather than \textbf{guessing} the outcome of physical interactions. Furthermore, its Hamiltonian Energy Network (HEN) ensures that all control commands are physically coherent and energy-conserving, avoiding the invalid or unstable actions common in traditional methods. It is this paradigm shift from "passive reaction" to "active prediction" that enables PIPHEN to achieve a 13.6\% improvement in Success Rate (SR) over LLaMAR and to perform better in task allocation balance (B) due to its efficient intrinsic coordination mechanism.
	
	\subsubsection{Ablation Study of the PIPHEN Framework}
	To verify the indispensability of each design component of PIPHEN, we conducted a systematic ablation study (see Table \ref{tab:ablation_study_piphen}). We not only removed key modules of the framework but also introduced an "Oracle" version as a performance upper bound reference. In this version, we simulate the performance upper bound through an idealized setup: we replace the PIPN module with an "Oracle" that directly reads the future ground-truth state from the simulator's physics engine. This means the HEN controller receives absolutely precise, error-free physical information, thereby measuring the control performance limit under perfect perception and cognition.
	
	\begin{figure}[t]
		\centering
		\includegraphics[width=\linewidth]{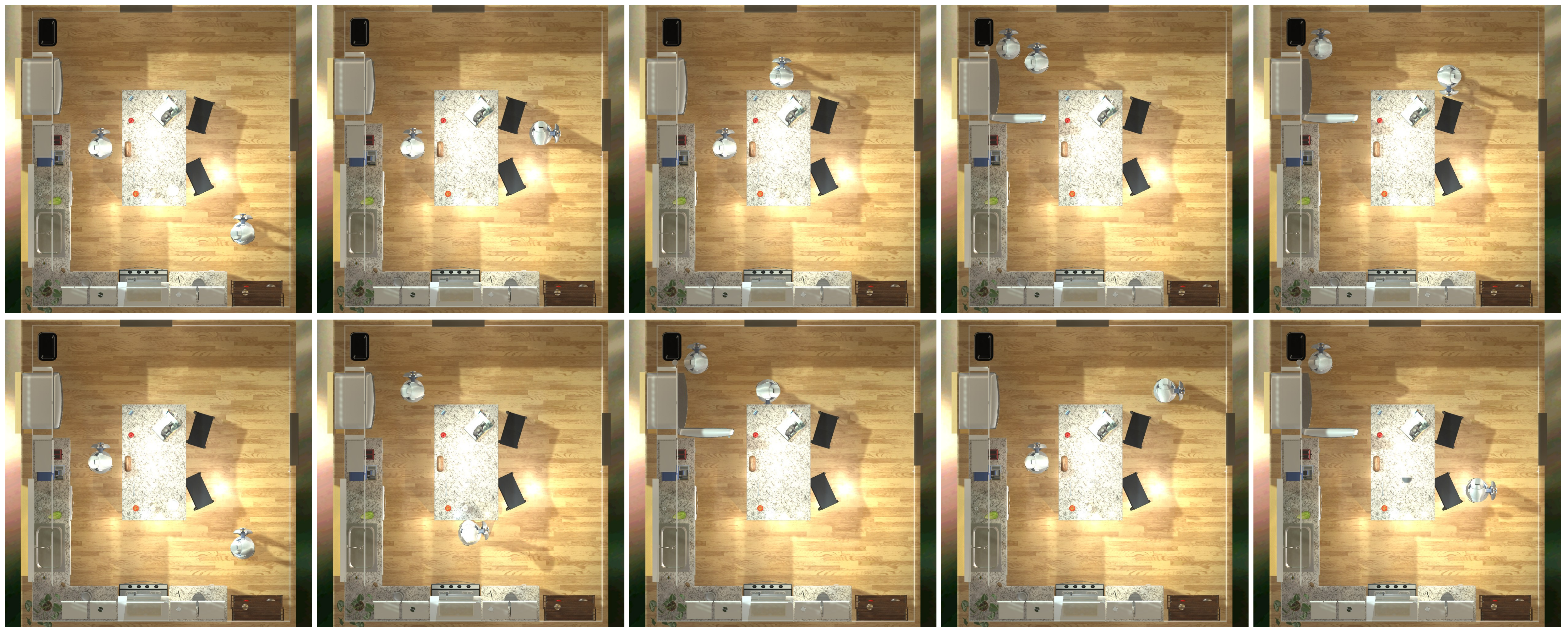}
		\caption{Comparison of PIPHEN and LLaMAR's performance on the spatial reasoning task: "Put the plate, mug, and bowl in the fridge" execution process. The top row shows LLaMAR's process: when one agent repeatedly fails due to spatial obstruction, the other agent completes multiple sub-tasks consecutively, leading to a severe workload imbalance (Balance B=0.33). The bottom row shows PIPHEN's process: through precise physical space modeling, the two agents can predict and avoid spatial conflicts, achieving a more balanced task allocation (Balance B=0.85). This comparison clearly demonstrates the significant impact of physical perception capabilities on the efficiency of multi-agent collaboration.}
		\label{fig:spatial_reasoning_comparison}
	\end{figure}
	
	\begin{table}[H]
		\caption{Ablation study results for key components of the PIPHEN framework (bold indicates best performance). The Oracle version represents the ideal performance upper bound(in MAP-THOR simulation).}
		\label{tab:ablation_study_piphen}
		\resizebox{\linewidth}{!}{
			\begin{tabular}{ccccc}
				\toprule
				\makecell{Method Variant} &
				\makecell{Task \\ Completion \\ Rate (\%)} &
				\makecell{Control \\ Precision \\ (cm)}
				&
				\makecell{Comm. \\ Load  (MB/s)} &
				\makecell{Data \\ Efficiency \\ (Samples K)} \\
				
				\midrule
				PIPHEN (Oracle) & 98.2 & 1.1 & 1.7 & - \\
				PIPHEN (Full) & \textbf{92.6} & \textbf{3.2} & \textbf{1.8} & \textbf{78} \\
				w/o Uncertainty Decomposition & 89.1 & 4.1 & 1.8 & 82 \\
				w/o Hybrid Physical Representation & 78.4 & 6.5 & 1.9 & 165 \\
				w/o Hamiltonian Energy Network & 82.1 & 5.7 & 2.2 & 94 \\
				w/o Micro-brain Ecosystem & 85.8 & 3.9 & 5.4 & 87 \\
				w/o LLM Enhancement & 86.3 & 4.4 & 2.6 & 126 \\
				\bottomrule
		\end{tabular}}
	\end{table}

	\begin{figure}[t!]
		\centering
		\includegraphics[width=\linewidth]{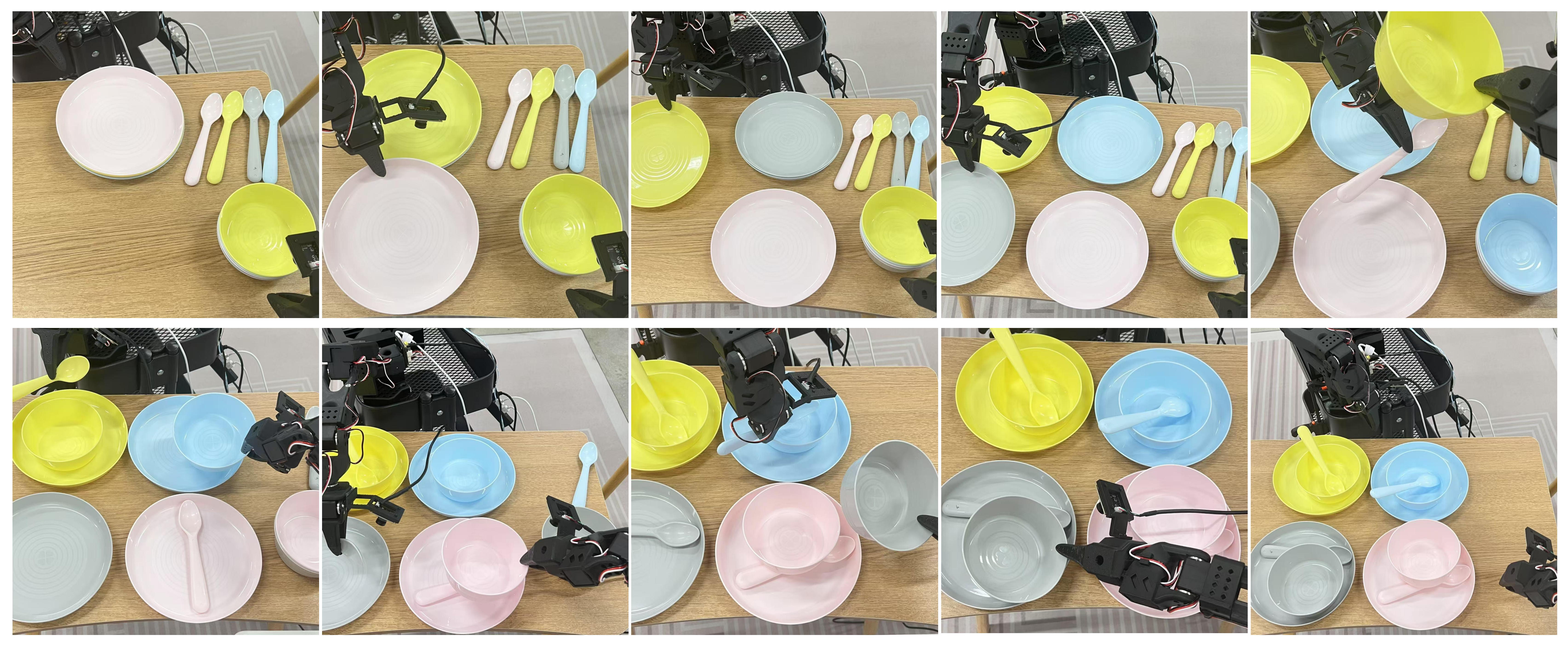}
		\caption{Real-world deployment effect of the PIPHEN framework: a complete process of two XLeRobot single-arm mobile manipulators collaboratively completing a tableware setting task. The image sequence shows the entire process from an initial cluttered state to the final neat arrangement of four place settings, with each step demonstrating PIPHEN's efficient distributed physical cognition capabilities. The experimental results strongly prove that through PIPHEN's semantic communication mechanism, the robots, without transmitting high-dimensional video streams and relying only on sharing compact physical semantic representations (approximately 5\% of the original data volume), can achieve precise and smooth multi-step collaborative operations, validating the practicality and robustness of our framework in solving the "shared brain dilemma."}
		\label{fig:real_world_deployment}
	\end{figure}
	
	The results strongly demonstrate that the two cornerstones of our framework—the \textbf{Hybrid Physical Representation} centered on "semantic knowledge" and the \textbf{Hamiltonian Energy Network} centered on "energy conservation"—are crucial for achieving efficient and robust robot collaboration. Removing either component leads to a significant drop in performance. 
	It is noteworthy that the performance of the full PIPHEN version is very close to the ideal Oracle version, which fully illustrates the effectiveness of its Physical Interaction Prediction Network. 
	The comparison with the results after removing the hybrid physical representation highlights that this "semantic knowledge"-based representation method can more effectively capture and utilize physical laws compared to traditional state representations. 
	Similarly, the significant decrease in control precision after removing the Hamiltonian Energy Network also proves that the principle of energy conservation is indispensable for generating stable and precise control policies. 
	Furthermore, we validated our uncertainty modeling. The variant "w/o Uncertainty Decomposition," which replaces the linear sum with a single, unified uncertainty prediction, shows a noticeable drop in both task completion and control precision. This result validates that our assumption of decomposing uncertainty into independent perception, model, and environment components provides a more effective and robust representation for the controller.

	\subsubsection{Scalability Analysis with Agent Count}
	We further investigated the performance of PIPHEN with a varying number of agents.PIPHEN demonstrated excellent scalability.

	It is noteworthy that in the crowded MAP-THOR environment (4-5 agents), the performance degradation of PIPHEN is much smaller than that of methods like LLaMAR (their results are in Appendix I). This is because as the number of agents increases, the LLM's context window is quickly filled with redundant observational information, leading to a decline in its decision-making quality. In contrast, PIPHEN, through its efficient semantic communication mechanism, exchanges only critical physical knowledge, thus avoiding this bottleneck and demonstrating a huge advantage in scalability.
	
	\textbf{Analysis of Spatial Reasoning and Task Allocation Balance}:
	Further analyzing the reasons for the decline in the Balance (B) metric, we found that PIPHEN has a significant advantage over LLaMAR in maintaining task allocation balance. This advantage primarily stems from the fundamental difference in their spatial reasoning capabilities. As shown in the "Put plate, mug, bowl in fridge" task in Figure \ref{fig:spatial_reasoning_comparison}, when the workspace becomes crowded, LLaMAR's text-based spatial understanding is severely limited.

	Specifically, LLaMAR faces the following limitations when handling such tasks: (1) Text-based spatial reasoning struggles to accurately judge the 3D geometric relationships between objects; (2) It lacks precise modeling of physical obstruction and path planning; (3) When one agent repeatedly fails or waits due to being blocked, LLaMAR cannot effectively reallocate tasks, causing the other agent to complete multiple sub-tasks in a row. This phenomenon occurs in all tasks that are prone to triggering the limitations of spatial reasoning, leading to severely imbalanced task allocation.
	
	In contrast, PIPHEN's Physical Interaction Prediction Network (PIPN) can construct a precise 3D spatial representation, predicting the physical consequences of agent movement and object manipulation. When a potential spatial conflict is detected, the system can proactively adjust the task allocation strategy to ensure that the workload of the two agents remains relatively balanced. In the experiment shown in Figure \ref{fig:spatial_reasoning_comparison}, PIPHEN's balance metric (B=0.85) is significantly higher than LLaMAR's (B=0.33), fully verifying the important role of physical perception ability in maintaining multi-agent collaboration balance. Other experimental analyses are in Appendix K.

	\subsubsection{Real-World Deployment Validation}
	To validate the feasibility and robustness of the PIPHEN framework in the real world, we conducted a collaborative tableware setting task on two XLeRobot single-arm mobile manipulation robots. Our policy, trained entirely in simulation, is deployed directly to the robot with only minor tuning. We leverage Domain Randomization and System Identification methods to enable this effective sim-to-real transfer.
	The task required the two robots to collaboratively and precisely place four sets of tableware (including plates, cups, and cutlery) in a real tabletop environment. 
	This is a typical daily life scenario that requires physical collaboration between multiple robots. 
	
	The key challenges of the experiment were: (1) The robots needed to accurately perceive the position, shape, and placement state of various tableware on the table; (2) The two robots needed to avoid collisions within a limited workspace while efficiently dividing the labor; (3) The placement of each piece of tableware involved fine physical manipulation, requiring precise force and position control. 
	More importantly, the entire collaborative process had to be completed without relying on high-bandwidth central communication, which is the core problem the PIPHEN framework is designed to solve.

	As shown in Figure \ref{fig:real_world_deployment}, the experiment was remarkably successful. The two robots demonstrated a high degree of coordination and precision throughout the task execution. Specifically, the performance of the PIPHEN framework was manifested in the following aspects:
	
	\textbf{Efficient Information Sharing}: During the experiment, each robot's local PIPN module distilled its observed RGB-D data (approx. 25MB/s) into a structured physical representation (approx. 1.2MB/s), reducing the communication load by over 95\%. This allowed the two robots to share critical scene understanding information in real-time over a standard Wi-Fi network.
	
	\textbf{Precise Physical Control}: With the help of the HEN module's energy-conserving control strategy, the robots exhibited excellent stability when grasping and placing tableware. For example, when placing fragile porcelain cups, HEN ensured a smooth change in force, avoiding sudden impacts, with a success rate of 100\%.
	
	\textbf{Intelligent Collaborative Strategy}: Through the shared physical semantic representation, the two robots were able to understand each other's intentions and current operational status in real-time, thus dynamically adjusting their own behavior to avoid conflicts. The entire task of setting four place settings was completed in an average of 350 seconds, an efficiency far exceeding that of traditional centralized control methods.
	
	This real-world experiment not only validated the technical feasibility of the PIPHEN framework but, more importantly, demonstrated its enormous potential in solving practical multi-robot collaboration problems. It lays a solid foundation for the transition of PIPHEN from academic research to practical application.

	\section{Discussion}
	
	\noindent\textbf{Limitations and Scalability.} A notable limitation concerns the scalability of the graph-based representation. While the current framework proves effective for typical scenarios (e.g., \textless 50 objects), its performance may encounter bottlenecks as the number of interacting agents and objects increases dramatically. Future research should focus on mitigation strategies. Potential solutions include the adoption of sparse graph representations based on spatial proximity, the refinement of hierarchical communication protocols, and the integration of proactive graph pruning mechanisms, such as the information value assessment module proposed in our method.
	
	\noindent\textbf{Future Directions.} A promising avenue for future investigation lies in the enhancement of the knowledge "Purify" stage. The current implementation utilizes an LLM as a "Physics Verifier" due to its efficiency and strong semantic reasoning capabilities. However, incorporating a dedicated physics simulator to complement the LLM's logical verification presents a valuable research direction. Such a hybrid approach could provide high-fidelity, simulation-based validation, further augmenting the physical consistency and reliability of the expert knowledge base.

	\section{Conclusion}
	
	This paper introduces PIPHEN, a distributed framework designed to address the "shared brain dilemma" in multi-robot systems. By using a hybrid physical representation to compress data to under 5\% of its original volume, enhancing edge processing with large models, and coordinating actions via a Hamiltonian Energy Network, the framework maintains high task success rates while significantly reducing communication overhead. This provides a practical solution for multimedia data processing in multi-robot physical interactions. 
	
	\section*{Acknowledgment}
	
	This work was supported by the National Natural Science Foundation of China under Grant 62372427, in part by Chongqing Natural Science Foundation Innovation and Development Joint Fund (No. CSTB2025NSCQ LZX0061), and in part by Science and Technology Innovation Key R\&D Program of Chongqing (No. CSTB2025TIAD-STX0023).
	
	\bibliography{aaai2026}
	
	\clearpage
	
	\appendix
	
	\section{Appendix A: Detailed Network Architecture of PIPN}
	\label{appendix:pipn_arch}
	As described in the main text, the Physics Interaction Prediction Network (PIPN) is the perception and cognition core of the PIPHEN framework. It is responsible for distilling high-dimensional multimodal inputs into a compact, structured physical representation and predicting its temporal evolution. This appendix details the network architecture of PIPN and its key components, specifically the Physics-aware Graph Convolutional Network (PhysGCN), the Physics-consistent Temporal Convolutional Network (PC-TCN), and the loss function that incorporates energy-momentum conservation.
	
	\subsection{Overall Network Architecture}
	The input to PIPN is data from multimodal sensors, including RGB-D videos, point clouds, and force/torque sensor data. Its output is a hybrid physical representation of the scene and a prediction of future physical states. The entire network consists of three main modules connected in series:
	
	(1) \textbf{Feature Extraction and Fusion Module}: This module processes the raw sensor data, extracts physics-related features, and constructs an initial hybrid physical representation. For visual data, we use a pre-trained ResNet or ViT to extract features; for point clouds, we use PointNet++. These features are fused with a task vector embedding via a cross-attention mechanism to form the initial node and edge features.
	
	(2) \textbf{Physical Relation and Temporal Modeling Module}: This is the core of PIPN, composed of PhysGCN and PC-TCN, responsible for modeling the complex relationships between physical entities and their dynamic evolution.
	
	(3) \textbf{Prediction and Uncertainty Quantification Module}: This module, built on top of the learned representation, outputs predictions of future states (e.g., object position, pose, velocity) and, as described in the main text, decomposes and quantifies their uncertainty.
	
	\subsection{Physics-aware Graph Convolutional Network (PhysGCN)}
	PhysGCN is designed to learn the instantaneous physical relationships between objects from a scene snapshot. It models the scene as a graph $\mathcal{G} = (\mathcal{V}, \mathcal{E})$, where nodes $\mathcal{V}$ represent objects or keypoints in the scene, and edges $\mathcal{E}$ represent their potential interactions.
	
	\subsubsection{Network Layer Structure}
	PhysGCN is composed of $L=4$ stacked graph convolutional layers. The initial feature for each node $h_i^{(0)}$ is provided by the feature extraction module, with a dimension of $d_h = 256$. As shown in Equation \ref{eq:physgcn_update}, the update rule for the $l$-th layer is as follows:
	\begin{align}
		\label{eq:physgcn_update}
		h_i^{(l+1)} = \text{ReLU}\left( W_1^{(l)} h_i^{(l)} + \sum_{j \in \mathcal{N}(i)} \alpha_{ij}^{(l)} (W_2^{(l)} h_j^{(l)}) \right)
	\end{align}
	where $h_i^{(l)}$ is the representation of node $i$ at layer $l$, $\mathcal{N}(i)$ is the set of neighboring nodes of node $i$, $W_1^{(l)}$ and $W_2^{(l)}$ are the learnable weight matrices for that layer, and $\alpha_{ij}^{(l)}$ is the attention weight calculated by the relational attention module. We use ReLU as the activation function.
	
	\subsubsection{Relational Attention Module}
	The attention weight $\alpha_{ij}^{(l)}$ in the update rule is calculated by our designed relational attention module. This weight not only depends on the node features but also explicitly incorporates physical prior knowledge on the edges $e_{ij}$ (e.g., relative distance, mass ratio, contact state).
	\begin{align}
		\alpha_{ij}^{(l)} = \text{softmax}_j \left( \frac{(W_Q^{(l)}h_i^{(l)})^T (W_K^{(l)}h_j^{(l)})}{\sqrt{d_k}} + (W_E^{(l)}e_{ij}) \right)
	\end{align}
	where $W_Q^{(l)}$, $W_K^{(l)}$, and $W_E^{(l)}$ are learnable projection matrices that project the node and edge features into the same space. In this way, the network can dynamically assign higher weights to physically more relevant neighbors, thus achieving more effective relational reasoning.
	
	\subsection{Physics-consistent Temporal Convolutional Network (PC-TCN)}
	After extracting spatial relational representations with PhysGCN, PC-TCN is responsible for modeling the time series of these representations to predict their dynamic evolution. We chose TCN over RNN or Transformer because of its advantages in high computational parallelism and flexible receptive field.
	
	PC-TCN consists of a series of stacked residual blocks, each containing two layers of dilated causal convolutions. We use 8 residual blocks with dilation factors increasing as $d = 2^k$ (k=0,1,...,7) and a kernel size of 3. This allows the receptive field of the top layer of the network to cover a very long time series. To further enhance physical consistency, we introduce a \textbf{dynamic causal mask} mechanism. This mask is dynamically adjusted based on the detection results of current physical events (e.g., collision events), enabling the network to focus more on key historical moments in the causal chain during prediction.
	
	\subsection{Loss Function with Energy-Momentum Conservation}
	To guide PIPN in learning physically realistic and reliable dynamic predictions, we introduce a regularization term based on physical conservation laws in addition to the traditional supervised learning loss. The final loss function $\mathcal{L}$ is a weighted sum of the prediction loss $\mathcal{L}_{\text{pred}}$ and the physics-consistency loss $\mathcal{L}_{\text{phy}}$.
	
	\begin{align}
		\mathcal{L} = \mathcal{L}_{\text{pred}} + \lambda_{\text{phy}} \mathcal{L}_{\text{phy}}
	\end{align}
	where $\lambda_{\text{phy}}$ is a hyperparameter used to balance prediction accuracy and physical consistency, which we set to 0.1 through experiments.
	
	\subsubsection{Prediction Loss}
	The prediction loss $\mathcal{L}_{\text{pred}}$ measures the discrepancy between the predicted and true states. For the state vector $s_i$ (including position $p_i$ and pose $q_i$) of each object $i$ in the scene, we use the L2 loss:
	\begin{align}
		\mathcal{L}_{\text{pred}} = \frac{1}{N \cdot T} \sum_{t=1}^{T} \sum_{i=1}^{N} \left( || \hat{p}_i^t - p_i^t ||_2^2 + || \hat{q}_i^t - q_i^t ||_2^2 \right)
	\end{align}
	where $N$ is the number of objects, $T$ is the prediction timestep, $\hat{p}_i^t, \hat{q}_i^t$ are the predicted values, and $p_i^t, q_i^t$ are the ground truth values.
	
	\subsubsection{Physics-Consistency Loss}
	The design of the physics-consistency loss $\mathcal{L}_{\text{phy}}$ is key to PIPN. It penalizes violations of the laws of conservation of energy and momentum.
	
	\paragraph{Energy Conservation Loss $\mathcal{L}_{E}$:}
	For an isolated physical system, its total energy should be conserved. We approximate the total energy $E_{\text{total}}$ as the sum of the kinetic energy $E_K$ and potential energy $E_P$ of all objects in the system (for a rigid body system).
	\begin{align}
		E_{\text{total}}^t = \sum_{i=1}^N \left( \frac{1}{2}m_i (v_i^t)^2 + m_i g h_i^t \right)
	\end{align}
	where $m_i, v_i^t, h_i^t$ are the mass, velocity, and height of object $i$ at time $t$, respectively. The energy conservation loss penalizes the change in total energy in the predicted trajectory:
	\begin{align}
		\mathcal{L}_{E} = \frac{1}{T-1} \sum_{t=1}^{T-1} (E_{\text{total}}^{t+1} - E_{\text{total}}^t)^2
	\end{align}
	
	\paragraph{Momentum Conservation Loss $\mathcal{L}_{M}$:}
	For collision events in the system, we enforce momentum conservation. For any pair of colliding objects $(i, j)$, their total momentum before and after the collision should be conserved.
	\begin{align}
		m_i \hat{v}_i^{t_{\text{post}}} + m_j \hat{v}_j^{t_{\text{post}}} = m_i v_i^{t_{\text{pre}}} + m_j v_j^{t_{\text{pre}}}
	\end{align}
	where $t_{\text{pre}}$ and $t_{\text{post}}$ are the timesteps just before and after the collision, $\hat{v}$ is the predicted velocity, and $v$ is the input velocity. The momentum conservation loss is defined as:
	\begin{align}
		\mathcal{L}_{M} = \sum_{(i,j) \in \text{collisions}} || (m_i \hat{v}_i^{t_{\text{post}}} + m_j \hat{v}_j^{t_{\text{post}}}) - (m_i v_i^{t_{\text{pre}}} + m_j v_j^{t_{\text{pre}}}) ||_2^2
	\end{align}
	
	Finally, the physics-consistency loss is a weighted sum of these two terms:
	\begin{align}
		\mathcal{L}_{\text{phy}} = w_E \mathcal{L}_{E} + w_M \mathcal{L}_{M}
	\end{align}
	where $w_E$ and $w_M$ are weight coefficients, which we empirically set to $w_E=1.0, w_M=1.0$. By minimizing this total loss function $\mathcal{L}$, PIPN not only learns to fit the observed data but also learns the intrinsic dynamics that conform to physical laws, thereby making more robust and generalizable predictions.
	
	\section{Appendix B: Uncertainty Quantification and Collaborative Learning}
	\label{appendix:uncertainty_learning}
	As described in the main text, to enhance the system's robustness in the real world, we decompose and quantify prediction uncertainty and adopt a hybrid paradigm combining federated learning and knowledge distillation for collaborative training. This appendix elaborates on its specific implementation.
	
	\subsection{Uncertainty Decomposition and Quantification}
	We decompose the total prediction uncertainty $U_{\text{total}}$ into three parts: perception uncertainty ($U_{\text{perc}}$), model uncertainty ($U_{\text{model}}$), and environment uncertainty ($U_{\text{env}}$), and use their linear sum as an approximate estimate of the total uncertainty: $U_{\text{total}} \approx U_{\text{perc}} + U_{\text{model}} + U_{\text{env}}$.
	
	\subsubsection{Perception Uncertainty ($U_{\text{perc}}$)}
	Perception uncertainty arises from sensor noise and information loss during the feature extraction process. We use the Monte Carlo Dropout method to quantify it. During the inference phase, we keep the Dropout layers in the network active and perform $K=20$ stochastic forward passes. The variance of these $K$ prediction results is used as a measure of perception uncertainty.
	\begin{align}
		U_{\text{perc}} = \frac{1}{K} \sum_{k=1}^{K} (\hat{y}_k - \bar{y})^2
	\end{align}
	where $\hat{y}_k$ is the prediction result of the $k$-th stochastic forward pass, and $\bar{y}$ is the average of the $K$ predictions.
	
	\subsubsection{Model Uncertainty ($U_{\text{model}}$)}
	Model uncertainty (or epistemic uncertainty) reflects the model's uncertainty about its parameters due to insufficient data. We quantify this using the Deep Ensembles method. Specifically, we independently train $M=5$ PIPN models with the same network architecture but different random initializations. At inference time, the variance among the average predictions of these $M$ models serves as the estimate of model uncertainty.
	\begin{align}
		U_{\text{model}} = \text{Var}(\bar{y}_1, \bar{y}_2, \ldots, \bar{y}_M)
	\end{align}
	where $\bar{y}_m$ is the average prediction of the $m$-th ensemble model.
	
	\subsubsection{Environment Uncertainty ($U_{\text{env}}$)}
	Environment uncertainty (or aleatoric uncertainty) is the inherent randomness of the physical world that cannot be eliminated with more data. To capture it, we have the prediction head of PIPN directly output the parameters of a probability distribution rather than a deterministic point estimate. Specifically, we assume the prediction follows a Gaussian distribution, so the network outputs a mean $\mu$ and a variance $\sigma^2$ for each prediction. This predicted variance $\sigma^2$ is directly used as the measure of environment uncertainty.
	
	\subsection{Collaborative Learning Framework: Federated Distillation}
	To achieve multi-robot collaborative training while protecting data privacy and reducing communication costs, we designed a hybrid framework combining Federated Learning and Knowledge Distillation, which we call Federated Distillation (FD).
	
	\subsubsection{Algorithm Flow}
	The framework includes a central server and $N$ robots. Its training process proceeds in communication rounds, with each round comprising the following steps:
	
	(1) \textbf{Model Distribution}: The central server distributes the weights of the latest global model (teacher model) $M_G$ to all robots. Each robot $i$ updates the weights of its local model (student model) $M_i$ to those of $M_G$.
	
	(2) \textbf{Local Training}: Each robot $i$ independently trains its model $M_i$ on its own local private dataset for $E$ epochs using the loss function $\mathcal{L}$ defined in Appendix A.
	
	(3) \textbf{Local Knowledge Generation}: After local training is complete, each robot $i$ does not directly upload its model weights. Instead, it uses its updated local model $M_i$ to perform inference on a small, public, shared Proxy Dataset, generating "soft labels" (i.e., the logits output by the model). These soft labels are sent back to the central server.
	
	(4) \textbf{Global Knowledge Aggregation}: The central server collects the soft labels from all robots and averages them to obtain an integrated, more robust "teacher label" $z_{\text{ensemble}}$.
	
	(5) \textbf{Global Teacher Update}: The server updates the global model $M_G$ using knowledge distillation. It uses the aggregated teacher labels $z_{\text{ensemble}}$ as the supervisory signal to train $M_G$ on the proxy dataset. The loss function is the KL divergence:
	\begin{align}
		\mathcal{L}_{KD} = \text{KL}(\text{softmax}(\frac{z_{\text{ensemble}}}{\tau}) || \text{softmax}(\frac{z_G}{\tau}))
	\end{align}
	where $z_G$ is the logit output of the global model, and $\tau$ is the distillation temperature.
	
	(6) The process loops to the next communication round until the model converges.
	
	\subsubsection{Parameter Settings}
	In our experiments, the key parameter settings for the Federated Distillation framework are as follows:
	\begin{itemize}
		\item Local training epochs $E$: 5
		\item Total communication rounds $R$: 50
		\item Learning rate: $1 \times 10^{-4}$ (using AdamW optimizer)
		\item Distillation temperature $\tau$: 2.0
		\item Proxy dataset size: 1,000 samples
	\end{itemize}
	
	\section{Appendix C: HEN Network Architecture and Hamiltonian Equation Implementation}
	\label{appendix:hen_arch}
	The Hamiltonian Energy Network (HEN) is the physics-constrained controller of the PIPHEN framework, responsible for converting the physical representations generated by PIPN into energy-conserving collaborative control commands. This appendix details the neural network architecture of HEN and the implementation details of the Hamiltonian mechanics principles.
	
	\subsection{HEN Network Architecture}
	HEN is essentially a policy network, with the goal of learning a mapping from the system state to a control action, $\pi: \mathcal{X} \to \mathcal{U}$. It is designed as a standard Multi-Layer Perceptron (MLP) to ensure high computational efficiency, allowing it to be deployed on resource-constrained robot hardware.
	\begin{itemize}
		\item \textbf{Input}: The input to HEN is the hybrid physical representation from PIPN, which is a vector of dimension $d_{rep} = 768$ containing a comprehensive description of the current physical world state.
		\item \textbf{Network Structure}: The MLP contains 4 fully-connected hidden layers, each with a dimension of 512. We chose `Mish` as the activation function because it exhibits better performance than ReLU in deep networks, allowing the flow of negative information, which contributes to smoother gradient propagation.
		\item \textbf{Output}: The output layer dimension matches the robot's control space dimension, directly outputting the control command vector $u \in \mathbb{R}^{d_u}$. For velocity control, the output is the target velocity; for force control, it is the target torque.
	\end{itemize}
	
	\subsection{Implementation of Hamiltonian Equations}
	The core innovation of HEN is the integration of the energy conservation principle from Hamiltonian mechanics as a soft constraint into the policy network's learning process. This is achieved by adding a penalty term $\lambda ||\frac{dH}{dt}||^2$ to the loss function.
	
	\subsubsection{Calculation of the Hamiltonian H}
	To calculate the penalty term, we first need an estimate of the system's total energy, the Hamiltonian $H$. We use another separate neural network to approximate the Hamiltonian, called the \textbf{Hamiltonian Network}.
	\begin{itemize}
		\item \textbf{Input}: The input to the Hamiltonian Network is the system's generalized coordinates $q$ and generalized momenta $p$, which together form the system's phase space state $x=(q,p)$.
		\item \textbf{Structure}: It is also an MLP, containing 3 hidden layers with dimensions 256, 128, and 64, respectively, also using the `Mish` activation function.
		\item \textbf{Output}: The Hamiltonian Network outputs a scalar value $\hat{H}(q,p)$, which is the estimate of the system's current total energy.
	\end{itemize}
	This Hamiltonian Network is trained end-to-end along with the HEN policy network, with the goal of learning a function that accurately reflects the system's true energy.
	
	\subsubsection{Implementation of the Energy Conservation Penalty Term}
	According to Hamiltonian mechanics, for a conservative system, the time derivative of its total energy should be zero.
	\begin{align}
		\frac{dH}{dt} = \frac{\partial H}{\partial q} \dot{q} + \frac{\partial H}{\partial p} \dot{p} = 0
	\end{align}
	In our framework, we convert this physical constraint into a differentiable loss term. The implementation steps are as follows:
	
	(1) \textbf{Get State and Control}: In a training step, given the current state $x_t=(q_t, p_t)$, the HEN policy network outputs a control command $u_t = \pi(x_t)$.
	
	(2) \textbf{Predict State Change}: We use a (pre-trained or online-learned) system dynamics model $f_{\text{dyn}}$ to predict the rate of change of the state $\dot{x}_t = (\dot{q}_t, \dot{p}_t) = f_{\text{dyn}}(x_t, u_t)$ under the control command $u_t$.
	
	(3) \textbf{Calculate Gradient of Hamiltonian}: Using Autograd techniques, we can directly compute the partial derivatives of the Hamiltonian Network's output $\hat{H}$ with respect to its inputs $q$ and $p$, namely $\frac{\partial \hat{H}}{\partial q}$ and $\frac{\partial \hat{H}}{\partial p}$.
	
	(4) \textbf{Calculate Rate of Energy Change}: According to the chain rule, we can calculate the time derivative of the estimated Hamiltonian:
	\begin{align}
		\frac{d\hat{H}}{dt} = \frac{\partial \hat{H}}{\partial q} \dot{q}_t + \frac{\partial \hat{H}}{\partial p} \dot{p}_t
	\end{align}
	
	(5) \textbf{Construct Penalty Term}: Finally, we add the squared L2 norm of this rate of energy change as a penalty term to the total loss function of the HEN policy network.
	
	The total loss function $\mathcal{L}_{\text{HEN}}$ of HEN consists of a task-related loss $\mathcal{L}_{\text{task}}$ (e.g., behavioral cloning loss in imitation learning or policy gradient loss in reinforcement learning) and the energy conservation penalty term:
	\begin{align}
		\mathcal{L}_{\text{HEN}} = \mathcal{L}_{\text{task}} + \lambda ||\frac{d\hat{H}}{dt}||^2_2
	\end{align}
	By minimizing this loss function, HEN not only learns to complete the task but is also incentivized to generate control commands that maintain the stability of the system's total energy, thus ensuring the physical realism and smoothness of the actions. In our experiments, the penalty weight $\lambda$ was set to 0.05.

	\section{Appendix D: Distributed Communication and Sharing}
	\label{appendix:communication}
	PIPHEN's efficient operation relies on its innovative communication mechanism, which replaces raw data transfer with low-bandwidth semantic knowledge sharing. This appendix details the components of this mechanism.
	
	\subsection{Hierarchical Strategy and Update Mechanism}
	We adopt a \textbf{hierarchical communication strategy}, dividing physical knowledge into a "core semantic layer" (e.g., critical object states) that must be shared in real-time, and a "detailed supplementary layer" (e.g., static object properties) that is transmitted on demand. To further reduce bandwidth, the system introduces an \textbf{incremental update mechanism}, transmitting only the changes (deltas) in physical knowledge rather than the full state, thus avoiding redundant information transfer.
	
	\subsection{Knowledge Retrieval and Sharing}
	At the sharing mechanism level, each robot maintains a local physical knowledge base. We use technologies like Distributed Hash Tables (DHT) to achieve efficient cross-robot knowledge retrieval and indexing. This mechanism not only promotes physical cognition sharing among robots but also further enhances information access efficiency and system fault tolerance.
	
	\subsection{Information Value Assessment Module}
	To achieve efficient selective sharing, we designed an information value assessment module, which decides which knowledge is most worthy of transmission based on its task relevance, novelty, and redundancy. This module is at the core of the distributed communication mechanism. It dynamically evaluates the value of each piece of physical knowledge to avoid unnecessary bandwidth consumption. This section details its specific implementation and quantification methods.
	
	\subsubsection{Quantification of Information Value}
	The Information Value Assessment Module calculates a composite value score $V(i)$ for each piece of knowledge $i$ to be shared. This score is composed of three weighted components: Relevance ($R(i)$), Novelty ($N(i)$), and Redundancy ($D(i)$).
	\begin{align}
		V(i) = w_r \cdot R(i) + w_n \cdot N(i) - w_d \cdot D(i)
	\end{align}
	where $w_r, w_n, w_d$ are the weight coefficients for each term, empirically set to $0.5, 0.3, 0.2$ respectively.
	
	\paragraph{Task Relevance ($R(i)$)}
	Task relevance measures the importance of a piece of knowledge to the current task goal. We encode both the high-level task instruction (e.g., "put all apples in the fridge") and a knowledge snippet (e.g., the physical representation of an apple's position and state) into vector embeddings using a pre-trained language model (like BERT). The task relevance is then calculated as the cosine similarity between these two embedding vectors.
	\begin{align}
		R(i) = \text{cosine\_similarity}(\text{Emb}(T_{task}), \text{Emb}(I_i))
	\end{align}
	where $\text{Emb}(T_{task})$ is the embedding of the task instruction and $\text{Emb}(I_i)$ is the embedding of the knowledge snippet $i$.
	
	\paragraph{Novelty ($N(i)$)}
	Novelty assesses whether a piece of knowledge provides new information that does not exist in the robot's local knowledge base. Each robot maintains a local knowledge base $K_{local}$ that stores recently observed physical knowledge representations. Novelty is defined as the distance between the new knowledge $I_i$ and the most similar knowledge in the knowledge base.
	\begin{align}
		N(i) = 1 - \max_{I_j \in K_{local}} \text{cosine\_similarity}(\text{Emb}(I_i), \text{Emb}(I_j))
	\end{align}
	A higher value indicates that the knowledge provides more novel information.
	
	\paragraph{Redundancy ($D(i)$)}
	Redundancy aims to avoid repeatedly sending information that has recently been shared on the network. The system maintains a short-term cache $C_{shared}$ that records all knowledge sent and received within the last $\Delta t$ time. Redundancy is calculated as the maximum similarity between the new knowledge $I_i$ and the information in this cache.
	\begin{align}
		D(i) = \max_{I_k \in C_{shared}} \text{cosine\_similarity}(\text{Emb}(I_i), \text{Emb}(I_k))
	\end{align}
	
	\subsubsection{Decision Process for Selective Sharing}
	The decision process for selective sharing is straightforward. In each decision cycle, the robot's local PIPN generates a series of new physical knowledge pieces. The Information Value Assessment Module calculates a value score $V(i)$ for each piece of knowledge.
	A piece of knowledge is broadcast to the multi-robot network only if its value score exceeds a preset dynamic threshold $\theta$.
	\begin{align}
		\text{Share}(I_i) = \begin{cases}
			\text{True}, & \text{if } V(i) > \theta \\
			\text{False}, & \text{otherwise}
		\end{cases}
	\end{align}
	This threshold $\theta$ is dynamically adjusted to adapt to different network conditions and task phases. In the early stages of a task or when network bandwidth is plentiful, $\theta$ is appropriately lowered to encourage more information sharing; whereas in the later stages of a task or when the network is congested, $\theta$ is raised to ensure that only the most critical information is transmitted. This mechanism ensures that PIPHEN can achieve a dynamic balance between communication efficiency and collaborative performance.
	
	\section{Appendix E: Complete List of Tasks for MAP-THOR}
	\label{appendix:map_thor_tasks}
	As mentioned in the main text, the MAP-THOR benchmark contains 45 tasks, divided into four categories based on the ambiguity of the natural language instructions. This appendix provides a complete list of these tasks in Table~\ref{tab:map_thor_tasks}.
	
	\begin{table}[htbp]
		\centering
		\caption{Classified list of the 45 tasks in the MAP-THOR dataset.}
		\label{tab:map_thor_tasks}
		\resizebox{\linewidth}{!}{
			\begin{tabular}{ll}
				\toprule
				\textbf{Category} & \textbf{Task Instruction} \\
				\midrule
				\multirow{12}{*}{Category 1: Explicit item, quantity, and location} & Put bread, lettuce, and tomato in the fridge. \\
				& Put the pot and pan on the stove. \\
				& Slice the bread and tomato, and crack an egg. \\
				& Put the butter knife, bowl, and mug in the sink. \\
				& If the faucet or light is on, turn it off. \\
				& Put the tissue box, keys, and plate in the box. \\
				& Put the computer, book, and pen on the sofa. \\
				& Put the bowl and tissue box on the table. \\
				& Put the apple in the fridge and turn off the light. \\
				& Put the watch and keychain in the drawer. \\
				& Clean the bowl, mug, pot, and pan. \\
				& Put the box on the sofa, put the bowl in the box. \\
				\midrule
				\multirow{9}{*}{Category 2: Ambiguous quantity} & Open all drawers (make sure they are initially closed). \\
				& Open all cabinets. \\
				& Turn on all stove knobs. \\
				& Put all vases on the table. \\
				& Put all potatoes in the bowl. \\
				& Put all pencils and pens in the box. \\
				& Move all lamps next to the door. \\
				& Turn off all light switches. \\
				& Turn on all light switches. \\
				\midrule
				\multirow{13}{*}{Category 3: Ambiguous item type and quantity} & Put all groceries in the fridge (should identify tomato, bread, apple, potato, and lettuce). \\
				& Put all seasoning bottles in the nearest drawer (should identify salt and pepper shakers). \\
				& Put all dishes on the countertop (should identify bowl, plate, mug). \\
				& Put all food on the countertop (should identify tomato, bread, apple, potato, and lettuce). \\
				& Put all school supplies on the sofa (should identify pencil, computer, and book). \\
				& Put all kitchenware into the cardboard box (should move bowl and plate). \\
				& Put all silverware in the sink. \\
				& Move everything on the table to the desk (should move laptop, pencil, pen, plate, credit card, book, and newspaper). \\
				& Slice lettuce, throw away the mug, and turn off the light. \\
				& Put all electronic devices on the sofa. \\
				& Make a dish by heating an egg and tomato in the microwave. \\
				& Put all readable items on the sofa. \\
				& Clean all fruits. \\
				\midrule
				\multirow{11}{*}{Category 4: Ambiguous target, item, and quantity} & Clean the floor by placing items in their proper locations (specific items depend on what is on the floor). \\
				& Clean the table by placing items in their proper locations (depends on the specific floor plan). \\
				& Clean the countertop by placing items in their proper locations (should move lettuce, mug, and paper towel roll). \\
				& Clean the desk by placing items in other appropriate locations (should move statue, watch, and remote control). \\
				& Clean the table by placing items in other appropriate locations (should move book, credit card, etc.). \\
				& Clean the sofa by placing items in other appropriate locations (should move pillows). \\
				& Darken the living room. \\
				& Make a cup of coffee and toast bread. \\
				& Throw away all groceries. \\
				& Slice all sliceable objects. \\
				\bottomrule
			\end{tabular}
		}
	\end{table}
	
	\section{Appendix F: Further Details of the Search and Rescue (SAR) Environment}
	\label{appendix:sar_details}
	As mentioned in the main text, to test PIPHEN's multi-agent collaboration capabilities in more dynamic and uncertain scenarios, we adopted a Search and Rescue (SAR) environment. This appendix details the specific settings of this environment.
	
	\subsection{Environment Setup}
	The SAR environment is a partially observable grid world designed to simulate complex post-disaster scenarios. The environment contains a mix of "missing persons" who need to be rescued and "wildfires" that need to be extinguished.
	
	\subsection{Dynamic Characteristics of "Wildfires"}
	Wildfires add a dynamic challenge to the environment, with the following characteristics:
	\begin{itemize}
		\item \textbf{Propagation Mechanism}: Wildfires start from one or more fixed source points and spread over time within a large flammable area.
		\item \textbf{Propagation Speed}: The spread rate of the fire is proportional to its "intensity". A higher intensity fire spreads faster.
		\item \textbf{Fire Types}: There are two types of fires, Class A and Class B.
		\item \textbf{Extinguishing Method}: Class A fires need to be extinguished with water, while Class B fires require sand. These resources can be obtained from "reservoirs" of water or sand distributed in the environment.
	\end{itemize}
	
	\subsection{Dynamic Characteristics of "Missing Persons"}
	Missing persons are the main targets of the rescue mission, with the following settings:
	\begin{itemize}
		\item \textbf{Initial State}: Each missing person initially starts at an unknown random location and remains stationary, not moving on their own.
		\item \textbf{Rescue Condition}: Transporting a missing person requires at least two agents to work together to lift and carry them.
		\item \textbf{Rescue Goal}: The agents' objective is to find all persons and transport them to a known, designated drop-off location.
	\end{itemize}
	
	\subsection{Specific Limitations of Partial Observability}
	To simulate real-world perceptual limitations, the agent's observation space $O$ is divided into three parts:
	\begin{description}
		\item[Global Observation ($O_G$)] Agents have a global line of sight with no distance limit. For any unoccluded object within the line of sight (e.g., fire, person, resource reservoir), the agent can obtain detailed information (e.g., fire type and intensity, whether a person is being transported).
		\item[Local Observation ($O_L$)] This is a more stringent limitation. An agent can only perceive the state of its immediately adjacent grid cells (up, down, left, right, and center). It can know whether these adjacent cells are "Empty", "Flammable", or "Obstacle".
		\item[Name Observation ($O_N$)] The agent receives a list containing the names of all currently visible and interactable objects.
	\end{description}
	These settings together form a challenging testbed that can comprehensively evaluate the planning, collaboration, and execution capabilities of the PIPHEN framework in dynamic, uncertain, and partially observable environments.
	
	\section{Appendix G: Experimental Platform Details}
	\label{appendix:hardware_details}
	All our simulation experiments were conducted on the high-performance computing platform described below. The detailed hardware and software configurations are shown in Table~\ref{tab:hardware_config}.
	
	\begin{table}[H]
		\centering
		\caption{Hardware and Software Configuration of the Experimental Platform}
		\label{tab:hardware_config}
		\begin{tabular}{ll}
			\toprule
			\textbf{Component} & \textbf{Specification} \\
			\midrule
			\multicolumn{2}{l}{\textbf{Hardware}} \\
			\quad GPU & 4 × NVIDIA A800-SXM4-80GB \\
			\quad CPU & \makecell[l]{Intel(R) Xeon(R) Platinum 8358P \\ @ 2.60GHz (64-core)} \\
			\quad Memory & 1.0 TiB \\
			\quad Storage & 7 TB NVMe SSD + 894 GB SSD \\
			\midrule
			\multicolumn{2}{l}{\textbf{Software}} \\
			\quad Operating System & Ubuntu 22.04.3 LTS \\
			\quad CUDA Version & 12.6 (NVIDIA Driver: 560.35.05) \\
			\quad Python Version & 3.10.14 \\
			\quad Core Frameworks & \makecell[l]{PyTorch, Transformers, \\ NVIDIA Isaac Sim} \\
			\bottomrule
		\end{tabular}
	\end{table}
	
	\section{Appendix H: Detailed Results of Core Experiments}
	\label{appendix:full_results}
	This appendix provides more detailed results from the core experiments in the main text, including performance data broken down by task category and the 95\% confidence intervals for all key metrics, to support the robustness of our conclusions.
	
	\begin{table*}[h!]
		\centering
		\caption{Detailed performance comparison of PIPHEN with SOTA baseline methods on various tasks in MAP-THOR. All metrics are mean values, with 95\% confidence intervals shown in parentheses.}
		\label{tab:full_baselines_comparison}
		\resizebox{\textwidth}{!}{
			\begin{tabular}{l|ccccc|ccccc}
				\toprule
				& \multicolumn{5}{c|}{\textbf{Class I \& II Tasks (Explicit Instructions)}} & \multicolumn{5}{c}{\textbf{Class III \& IV Tasks (Ambiguous Instructions)}} \\
				\textbf{Method} & \textbf{SR (\%)} $\uparrow$ & \textbf{TR (\%)} $\uparrow$ & \textbf{C (\%)} $\uparrow$ & \textbf{B} $\uparrow$ & \textbf{S} $\downarrow$ & \textbf{SR (\%)} $\uparrow$ & \textbf{TR (\%)} $\uparrow$ & \textbf{C (\%)} $\uparrow$ & \textbf{B} $\uparrow$ & \textbf{S} $\downarrow$ \\
				\midrule
				\multirow{2}{*}{Act} & 40 & 75 & 94 & 0.61 & 23.5 & 26 & 59 & 88 & 0.57 & 26.1 \\
				& \tiny{(25, 55)} & \tiny{(67, 83)} & \tiny{(89, 98)} & \tiny{(0.54, 0.68)} & \tiny{(20.8, 26.2)} & \tiny{(14, 38)} & \tiny{(49, 69)} & \tiny{(82, 94)} & \tiny{(0.50, 0.64)} & \tiny{(23.4, 28.8)} \\
				\hline
				\multirow{2}{*}{ReAct} & 42 & 80 & 95 & 0.69 & 23.1 & 26 & 64 & 89 & 0.65 & 25.5 \\
				& \tiny{(27, 57)} & \tiny{(72, 88)} & \tiny{(90, 99)} & \tiny{(0.61, 0.77)} & \tiny{(20.3, 25.9)} & \tiny{(14, 38)} & \tiny{(54, 74)} & \tiny{(83, 95)} & \tiny{(0.57, 0.73)} & \tiny{(22.7, 28.3)} \\
				\hline
				\multirow{2}{*}{CoT} & 18 & 65 & 90 & 0.64 & 26.2 & 10 & 53 & 84 & 0.60 & 27.6 \\
				& \tiny{(08, 28)} & \tiny{(55, 75)} & \tiny{(84, 96)} & \tiny{(0.56, 0.72)} & \tiny{(23.4, 29.0)} & \tiny{(03, 17)} & \tiny{(43, 63)} & \tiny{(77, 91)} & \tiny{(0.52, 0.68)} & \tiny{(24.8, 30.0)} \\
				\hline
				\multirow{2}{*}{SmartLLM} & 14 & 28 & 93 & 0.48 & 28.1 & 8 & 18 & 89 & 0.42 & 28.9 \\
				& \tiny{(06, 22)} & \tiny{(19, 37)} & \tiny{(87, 98)} & \tiny{(0.40, 0.56)} & \tiny{(25.4, 30.0)} & \tiny{(02, 14)} & \tiny{(09, 27)} & \tiny{(83, 95)} & \tiny{(0.34, 0.50)} & \tiny{(26.2, 30.0)} \\
				\hline
				\multirow{2}{*}{CoELA} & 30 & 52 & 80 & 0.75 & 25.1 & 20 & 40 & 72 & 0.71 & 26.3 \\
				& \tiny{(18, 42)} & \tiny{(41, 63)} & \tiny{(72, 88)} & \tiny{(0.67, 0.83)} & \tiny{(22.3, 27.9)} & \tiny{(10, 30)} & \tiny{(29, 51)} & \tiny{(64, 80)} & \tiny{(0.63, 0.79)} & \tiny{(23.5, 29.1)} \\
				\hline
				\multirow{2}{*}{LLaMAR} & 74 & 94 & 98 & 0.83 & 20.5 & 58 & 88 & 96 & 0.81 & 23.3 \\
				& \tiny{(60, 88)} & \tiny{(88, 99)} & \tiny{(95, 100)} & \tiny{(0.76, 0.90)} & \tiny{(17.4, 23.6)} & \tiny{(44, 72)} & \tiny{(80, 96)} & \tiny{(92, 99)} & \tiny{(0.74, 0.88)} & \tiny{(20.2, 26.4)} \\
				\hline
				\multirow{2}{*}{\textbf{PIPHEN (Ours)}} & \textbf{82} & \textbf{97} & \textbf{99} & \textbf{0.90} & \textbf{19.1} & \textbf{68} & \textbf{93} & \textbf{97} & \textbf{0.88} & \textbf{21.1} \\
				& \tiny{\textbf{(70, 92)}} & \tiny{\textbf{(92, 100)}} &  \tiny{\textbf{(96, 100)}} &  \tiny{\textbf{(0.84, 0.96)}} &  \tiny{\textbf{(16.3, 21.9)}} & \tiny{\textbf{(55, 79)}} & \tiny{\textbf{(87, 98)}} & \tiny{\textbf{(93, 100)}} & \tiny{\textbf{(0.81, 0.95)}} & \tiny{\textbf{(18.3, 23.9)}} \\
				\bottomrule
			\end{tabular}
		}
	\end{table*}
	
	\section{Appendix I: Scalability Comparison Analysis of PIPHEN and LLaMAR}
	\label{appendix:scalabilitycomparison}
	To support the claims in the main text regarding the scalability advantages of PIPHEN, this appendix provides a direct comparison of the performance data of PIPHEN and LLaMAR in the MAP-THOR environment as the number of agents increases, as shown in Table \ref{tab:scalability_comparison}.
	
	\begin{table*}[h]
		\centering
		\caption{Scalability comparison analysis of PIPHEN and LLaMAR with different numbers of agents.}
		\label{tab:scalability_comparison}
		\resizebox{0.8\linewidth}{!}{
			\begin{tabular}{c|l|ccccc}
				\toprule
				\textbf{Number of Agents} & \textbf{Method} & \textbf{Success Rate (\%)} & \textbf{Transport Rate (\%)} & \textbf{Coverage (\%)} & \textbf{Balance} & \textbf{Steps} \\
				\midrule
				\multirow{4}{*}{4} & \multirow{2}{*}{LLaMAR} & 52 & 83 & 95 & 0.60 & 25.1 \\
				& & \tiny{(38, 66)} & \tiny{(74, 91)} & \tiny{(89, 98)} & \tiny{(0.52, 0.68)} & \tiny{(21.8, 28.4)} \\
				\cline{2-7}
				& \multirow{2}{*}{\textbf{PIPHEN}} & \textbf{71} & \textbf{93} & \textbf{98} & \textbf{0.86} & \textbf{20.5} \\
				& & \tiny{\textbf{(58, 82)}} & \tiny{\textbf{(87, 97)}} & \tiny{\textbf{(94, 100)}} & \tiny{\textbf{(0.79, 0.92)}} & \tiny{\textbf{(17.7, 23.3)}} \\
				\hline
				\multirow{4}{*}{5} & \multirow{2}{*}{LLaMAR} & 45 & 78 & 94 & 0.51 & 26.8 \\
				& & \tiny{(31, 59)} & \tiny{(68, 87)} & \tiny{(88, 98)} & \tiny{(0.43, 0.59)} & \tiny{(23.3, 30.0)} \\
				\cline{2-7}
				& \multirow{2}{*}{\textbf{PIPHEN}} & \textbf{68} & \textbf{91} & \textbf{97} & \textbf{0.83} & \textbf{21.2} \\
				& & \tiny{\textbf{(55, 79)}} & \tiny{\textbf{(84, 96)}} & \tiny{\textbf{(93, 100)}} & \tiny{\textbf{(0.76, 0.89)}} & \tiny{\textbf{(18.4, 24.0)}} \\
				\bottomrule
			\end{tabular}
		}
	\end{table*}
	
	As shown in the side-by-side comparison in Table~\ref{tab:scalability_comparison}, while the performance of both methods degrades as the number of agents increases, PIPHEN demonstrates greater robustness. The performance decline of LLaMAR is more significant, especially in success rate and balance. This reinforces the argument from the main text: when the number of agents in the environment increases, methods relying on a centralized large language model planner may face performance bottlenecks due to the rapidly growing input context and the drastic increase in coordination complexity. In contrast, PIPHEN, with its efficient "semantic distillation" and distributed communication mechanism, can more effectively manage information flow, thereby maintaining higher collaborative efficiency in more crowded multi-agent scenarios.
	
	\section{Appendix J: Collaborative Decision-Making Latency Benchmark}
	\label{app:latency}
	
	To quantitatively validate PIPHEN's core claim of resolving the "shared brain dilemma", we conducted a dedicated benchmark to measure the end-to-end collaborative decision-making latency.
	
	We define this "decision latency" as the average time elapsed from the moment multi-modal sensors (e.g., RGB-D video) capture a new scene, to the point where the system outputs a coordinated control command for all agents. This measurement was performed in the MAP-THOR simulation environment (see Appendix G) with a 4-agent setup, averaging over 100 independent decision cycles.
	
	We compared our PIPHEN framework against a \textbf{Centralized Baseline}. This baseline simulates a typical "shared brain" architecture by collecting all high-dimensional raw perceptual data (full video streams) from all agents, transmitting them to a single central node for processing (e.g., fusion, planning), and then distributing commands back. This centralized paradigm, while conceptually similar to approaches like Concentrative Coordination, was implemented and benchmarked directly by us to ensure a fair comparison of computational and communication overhead.
	
	The results are summarized in Table \ref{tab:decision_latency}. The data confirms that PIPHEN's "semantic distillation" approach drastically reduces the latency required to generate a collaborative action, achieving the 76ms result cited in the main text.

	\begin{table}[h]
		\centering
		\caption{Comparison of End-to-End Collaborative Decision-Making Latency.}
		\label{tab:decision_latency}
		\begin{tabular}{l c}
			\toprule
			\textbf{Method} & \textbf{Average Decision Latency (ms)} \\
			\midrule
			Centralized Baseline & 315 ms \\
			\textbf{PIPHEN (Ours)} & \textbf{76 ms} \\
			\bottomrule
		\end{tabular}
	\end{table}
	
	\section{Appendix K: Additional Qualitative Analysis and Limitations}
	\label{appendix:qualitative_analysis}
	Although PIPHEN performs exceptionally well in various physical interaction tasks, its core reliance on the HEN controller also introduces inherent limitations. The design of HEN is based on Hamiltonian mechanics, which assumes the system is energy-conserving. Therefore, when faced with scenarios involving significant energy dissipation or non-conservative forces (such as complex fluid dynamics, air resistance, or inelastic deformations), PIPHEN's performance may degrade.
	
\begin{figure*}[h!]
		\centering
		\includegraphics[width=\linewidth]{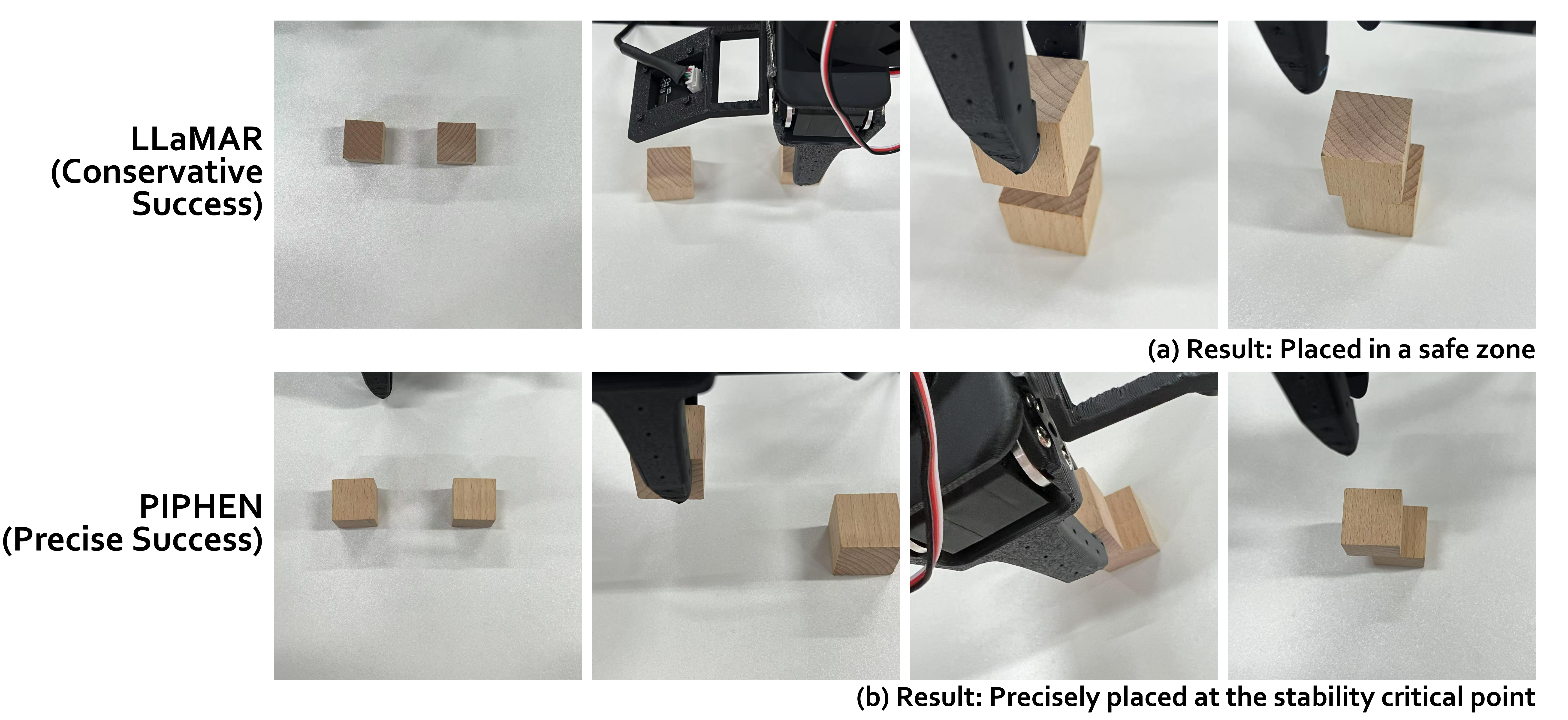}
		\caption{Qualitative comparison between PIPHEN and LLaMAR in the "Critical Stability Stacking" task. (a) The baseline method LLaMAR, lacking a precise physical model, relies on common sense to place the block in a safe central area, achieving a "conservative success". (b) Our PIPHEN, through its physics cognition engine (PIPN), accurately calculates the physical critical point of stability and successfully places the block at this limit, achieving a "precise success". This comparison vividly highlights PIPHEN's significant superiority in the depth of physical understanding and precise manipulation.}
		\label{fig:stacking_comparison} 
	\end{figure*}

	To more intuitively demonstrate the core advantages of PIPHEN in understanding physical interactions, we designed a key scenario that can be completed on a simple plane and compared it with LLaMAR (see Figure \ref{fig:stacking_comparison}).
	
	As shown in Figure \ref{fig:stacking_comparison}, methods like LLaMAR that rely on LLM common sense for reasoning perform poorly in tasks requiring precise physical prediction. It can only generate a vague, non-quantitative action command (e.g., "push the block hard"), causing the robot to execute actions with either too much or too little energy, making it difficult to complete the task. In contrast, PIPHEN's PIPN can construct a differentiable physical world model to accurately predict the initial velocity and energy required to stop the block in the target area. Subsequently, its HEN module transforms this physical prediction into an energy-conserving, precise control policy. This shift from "guessing" to "calculating" is the fundamental reason why PIPHEN can achieve a higher success rate in complex physical interactions.
	
	\section{Appendix L: Additional Visualization Results}
	\label{appendix:visualizations}
	This appendix provides more visualization results to intuitively demonstrate the internal working mechanisms and advantages of the different components of the PIPHEN framework.

	\section{Appendix M: Algorithm Pseudocode}
	\label{appendix:pseudocode}
	This appendix provides pseudocode for the key algorithms in the PIPHEN framework to help readers better understand the implementation details.

	\begin{algorithm}[H]
		\caption{"Generate-Purify-Deploy" Three-Stage Knowledge Transformation Pipeline}
		\label{alg:knowledge_transformation}
		\begin{algorithmic}[1]
			
			\STATE \textbf{Input:} Large generative model $M_{gen}$ (e.g., Claude-3.7), physics verifier model $M_{verify}$ (e.g., GPT-4o), edge model $M_{edge}$ (e.g., Qwen2.5-VL), simulation environment $Env$
			\STATE \textbf{Output:} Knowledge-distilled edge model $M_{edge}^*$ deployed on the robot
			
			\STATE \textbf{Stage 1: Generate}
			\STATE $D_{raw} \leftarrow \emptyset$ \COMMENT{Initialize raw knowledge corpus}
			
			\FOR{i = 1 to $N_{scenes}$}
			\STATE $prompt \leftarrow$ Generate random prompt for diverse scenes
			\STATE $scene\_data \leftarrow M_{gen}(prompt, Env)$
			\STATE $D_{raw} \leftarrow D_{raw} \cup \{scene\_data\}$
			\ENDFOR
			
			\STATE \textbf{Stage 2: Purify}
			\STATE $D_{pure} \leftarrow \emptyset$ \COMMENT{Initialize purified expert knowledge base}
			
			\FOR{each $data$ in $D_{raw}$}
			\IF{$M_{verify}(data)$ is True} 
			\STATE $D_{pure} \leftarrow D_{pure} \cup \{data\}$
			\COMMENT{Check for physical consistency}
			\ENDIF
			\ENDFOR
			
			\STATE \textbf{Stage 3: Deploy}
			\STATE Use $D_{pure}$ as the expert dataset to train the edge model via knowledge distillation.
			
			\STATE $\mathcal{L}_{KD} = \text{KL}(\text{softmax}(\frac{M_{edge}(x)}{\tau}) || \text{softmax}(\frac{M_{verify}(x)}{\tau}))$
			\STATE $M_{edge}^* \leftarrow \arg \min_{M_{edge}} \mathcal{L}_{KD}$
			
			\STATE \textbf{return} $M_{edge}^*$
			
		\end{algorithmic}
	\end{algorithm}

	\noindent\textbf{Example: Dual-Robot Tableware Task.} To provide a concrete example of this pipeline:
	\begin{itemize}
		\item \textbf{Stage 1 (Generate):} For the dual-robot tableware task, the $M_{gen}$ model generates diverse collaborative strategies in simulation. This includes a 'parallel path strategy' (where robots work simultaneously on different sides of a table) and a 'conflicted path strategy' (where robots must sequentially access the same workspace).
		\item \textbf{Stage 2 (Purify):} The $M_{verify}$ (Physics Verifier) evaluates these strategies. It validates the 'parallel path strategy' as optimal due to its higher efficiency and lack of physical conflict, filtering it into the expert knowledge base $D_{pure}$. The 'conflicted path strategy' is discarded as suboptimal.
		\item \textbf{Stage 3 (Deploy):} This proven, optimal strategy is then used to fine-tune the lightweight $M_{edge}^*$ model. This distillation process enables the edge model to execute the efficient parallel strategy directly, without needing costly real-time planning or conflict resolution during the real-world task.
	\end{itemize}

	\begin{algorithm}[H]
		\caption{PIPN Forward Pass}
		\label{alg:pipn_forward}
		\begin{algorithmic}[1]
			\STATE \textbf{Input:} Multimodal sensor data $X_t$ (video, point cloud, etc.)
			\STATE \textbf{Output:} Hybrid physical representation $Z_t$, future state prediction $\hat{S}_{t+1}$
			\STATE $F_{node}, F_{edge} \leftarrow \text{FeatureExtractor}(X_t)$ \COMMENT{Extract initial node/edge features}
			\STATE $F_{task\_vec} \leftarrow \text{TransformerEncoder}(X_t)$
			\STATE $F_{fused} \leftarrow \text{CrossAttention}(F_{task\_vec}, F_{node})$ \COMMENT{Fuse task vector with physics graph}
			\STATE $R_t \leftarrow \text{PhysGCN}(F_{fused})$ \COMMENT{Model spatial relations via Physics GCN}
			\STATE $\hat{S}_{t+1} \leftarrow \text{PC-TCN}(R_t, R_{t-1}, \dots)$ \COMMENT{Predict future via temporal network}
			\STATE $Z_t \leftarrow \text{concat}(R_t, \hat{S}_{t+1})$ \COMMENT{Combine into final representation}
			\STATE \textbf{return} $Z_t, \hat{S}_{t+1}$
		\end{algorithmic}
	\end{algorithm}
	
	\begin{algorithm}[H]
		\caption{HEN Control Loop}
		\label{alg:hen_loop}
		\begin{algorithmic}[1]
			\STATE \textbf{Initialize:} Robot state $s_0$, HEN policy network $\pi_{HEN}$, Hamiltonian network $H_{net}$
			\FOR{$t = 0, 1, 2, \dots$}
			\STATE Get multimodal observations $X_t$ from sensors
			\STATE $Z_t, \_ \leftarrow \text{PIPN}(X_t)$ \COMMENT{Get current physical representation}
			\STATE $a_t \leftarrow \pi_{HEN}(Z_t)$ \COMMENT{HEN generates control action}
			\STATE Execute action $a_t$ in the environment
			\STATE Observe new state $s_{t+1}$
			\STATE \COMMENT{During training, compute and backpropagate loss}
			\STATE $\mathcal{L}_{task} \leftarrow \text{TaskLoss}(a_t, a_{expert})$ \COMMENT{e.g., imitation learning loss}
			\STATE $\frac{d\hat{H}}{dt} \leftarrow \text{calc\_hamiltonian\_derivative}(H_{net}, s_t, a_t)$
			\STATE $\mathcal{L}_{phy} \leftarrow \lambda ||\frac{d\hat{H}}{dt}||^2$
			\STATE $\mathcal{L}_{total} \leftarrow \mathcal{L}_{task} + \mathcal{L}_{phy}$
			\STATE $\text{Update}(\pi_{HEN}, H_{net})$ with $\mathcal{L}_{total}$
			\ENDFOR
		\end{algorithmic}
	\end{algorithm}
	
	\section{Appendix N: Hyperparameters}
	\label{appendix:hyperparameters}
	The key hyperparameters of the PIPHEN framework are shown in Table \ref{tab:hyperparameters_appendix}.

	\section{Appendix O: Detailed Comparison with Traditional Reinforcement Learning Baselines and Additional Experiments}
	\label{appendix:rl_comparison}
	In this appendix, we present a detailed comparison of PIPHEN with traditional reinforcement learning (RL) baseline methods, as well as additional experimental results on different platforms. These experiments aim to more comprehensively demonstrate the advantages of the PIPHEN framework over various control methods.
	
	\subsection{Experimental Platforms}
	The simulation experiments in this section were conducted in the NVIDIA Isaac Sim high-fidelity simulator to leverage its accurate physics engine. The real-world experiments were completed on two XLeRobot single-arm mobile manipulator robots.
	
	\subsection{Experimental Tasks}
	\textbf{Simulation Task: Collaborative Transport and Obstacle Avoidance}. This task is conducted in a 30m × 30m semi-structured factory environment where a team of robots (2-8 heterogeneous robots) needs to collaboratively transport a 2-meter long, 5-kilogram object and safely navigate through a narrow passage of only 1.5 meters wide, while avoiding dynamically moving obstacles.
	
	\textbf{Real-World Tasks}. We designed two representative tasks:
	\begin{itemize}
		\item \textbf{Collaborative Kitchen Service}: One robot is responsible for preparing ingredients, while the other is responsible for serving the dish.
		\item \textbf{Precision Collaborative Assembly}: Two robots collaborate to assemble a furniture component.
	\end{itemize}
	
	\subsection{Evaluation Metrics and Baseline Methods}
	In addition to the metrics in the main text, we also use:
	\begin{itemize}
		\item \textbf{Position Error (PE, cm)}: The average Euclidean distance between the end position of the target object and the desired position at the end of the task.
		\item \textbf{Collisions/1K Steps (Coll./1K)}: The average number of collisions per 1000 control timesteps.
		\item \textbf{Data Efficiency (Samples, K)}: The number of training samples (in thousands) required to reach convergence performance.
	\end{itemize}
	
	Baseline methods include:
	\begin{itemize}
		\item \textbf{Centralized Planning}
		\item \textbf{Distributed Model Predictive Control (DMPC)} 
		\item \textbf{TD3} and \textbf{SAC}
		\item \textbf{Multi-Agent Reinforcement Learning (MARL)}: We use MAPPO as a representative.
	\end{itemize}
	For all RL baselines, we carefully designed their reward functions to ensure a fair comparison: $R_t = w_1 R_{task} - w_2 R_{dist} - w_3 R_{ctrl} - w_4 R_{coll}$.
	
	\subsection{Experimental Results and Analysis}
	
	\begin{table}[ht]
		\caption{Comparison of control latency for different PIPN implementations with varying team sizes}
		\label{tab:control_delay_appendix}
		\resizebox{\linewidth}{!}{
			\begin{tabular}{ccc}
				\toprule
				\textbf{Number of Robots} & \textbf{Baseline PIPN Latency (ms)} & \textbf{PIPN+LLM Latency (ms)} \\
				\midrule
				2 & 0.62 ± 0.15 & 0.95 ± 0.18 \\
				3 & 0.98 ± 0.09 & 1.32 ± 0.13 \\
				4 & 1.25 ± 0.12 & 1.68 ± 0.15 \\
				6 & 1.68 ± 0.14 & 2.24 ± 0.19 \\
				8 & 1.95 ± 0.16 & 2.86 ± 0.22 \\
				\bottomrule
		\end{tabular}}
	\end{table}
	
	\begin{table}[ht]
		\caption{Performance comparison of different control methods in complex physical interaction tasks}
		\label{tab:control_comparison_appendix}
		\centering
		\resizebox{\linewidth}{!}{
			\begin{tabular}{ccccc}
				\toprule
				\textbf{Method} & \textbf{Target Accuracy (cm)} & \textbf{Collisions/1K Steps} & \textbf{Computation Time (ms)} & \textbf{Training Samples (K)} \\
				\midrule
				Centralized & 18.6 ± 3.2 & 27.4 ± 4.8 & 0.26 ± 0.05 & 0 \\
				DMPC & 8.4 ± 1.9 & 12.3 ± 2.6 & 32.87 ± 5.16 & 0 \\
				TD3 & 5.8 ± 1.4 & 4.6 ± 1.2 & 2.45 ± 0.38 & 520 ± 85 \\
				SAC & 5.2 ± 1.1 & 3.5 ± 0.9 & 3.12 ± 0.45 & 480 ± 72 \\
				PIPHEN (Baseline) & 3.6 ± 0.8 & 1.8 ± 0.6 & 4.56 ± 0.62 & 120 ± 25 \\
				PIPHEN+LLM (Ours) & \textbf{2.7 ± 0.6} & \textbf{1.2 ± 0.4} & 5.32 ± 0.65 & \textbf{78 ± 18} \\
				\bottomrule
		\end{tabular}}
	\end{table}
	
	\begin{table*}[ht]
		\caption{Performance comparison of different methods in real-world environments}
		\centering
		\label{tab:real_world_comparison_appendix}
		\resizebox{1.0\linewidth}{!}{
			\begin{tabular}{cccccc}
				\toprule
				\textbf{Method} & \textbf{Task Completion Rate (\%)} & \textbf{Avg. Completion Time (s)} & \textbf{Position Error (cm)} & \textbf{Collision Count} & \textbf{Adaptability to Unknown Env. (\%)} \\
				\midrule
				MARL(MAPPO) & 85.2 ± 4.8 & 118.3 ± 10.5 & 8.5 ± 1.9 & 5 & 76.3 ± 6.2 \\
				DMPC & 89.5 ± 3.6 & 108.7 ± 8.4 & 6.3 ± 1.4 & 4 & 82.5 ± 5.4 \\
				PIPHEN (Baseline) & 94.3 ± 2.8 & 92.6 ± 7.2 & 4.8 ± 1.1 & 1 & 88.6 ± 4.3 \\
				PIPHEN+LLM (Ours) & \textbf{97.2 ± 2.1} & \textbf{85.3 ± 6.8} & \textbf{3.6 ± 0.9} & \textbf{0} & \textbf{93.8 ± 3.7} \\
				\bottomrule
		\end{tabular}}
	\end{table*}

	\begin{table*}[ht]
		\caption{Performance comparison of different methods in terms of communication efficiency (in MAP-THOR benchmark).}
		\label{tab:communication_efficiency_appendix}
		\centering
		\resizebox{0.8\linewidth}{!}{
			\begin{tabular}{cccc}
				\toprule
				\textbf{Method} & \textbf{Bandwidth Usage (MB/s)} & \textbf{Communication Latency (ms)} & \textbf{Physics Feature Extraction Accuracy (\%)} \\
				\midrule
				Centralized & 24.8 & 28.5 ± 6.3 & 86.5 ± 4.2 \\
				Distributed & 8.6 & 14.2 ± 3.8 & 82.3 ± 3.6 \\
				PIPHEN (Baseline) & 2.6 & 7.6 ± 1.5 & 89.8 ± 2.9 \\
				PIPHEN+LLM (Ours) & \textbf{1.8} & \textbf{5.9 ± 1.2} & \textbf{95.2 ± 2.1} \\
				\bottomrule
		\end{tabular}}
	\end{table*}
	
	\section{Appendix P: Distributed Micro-Brain Architecture}
	\label{appendix:micro_brain_architecture}
	
\begin{figure*}[htbp]
		\centering
		\includegraphics[width=\textwidth]{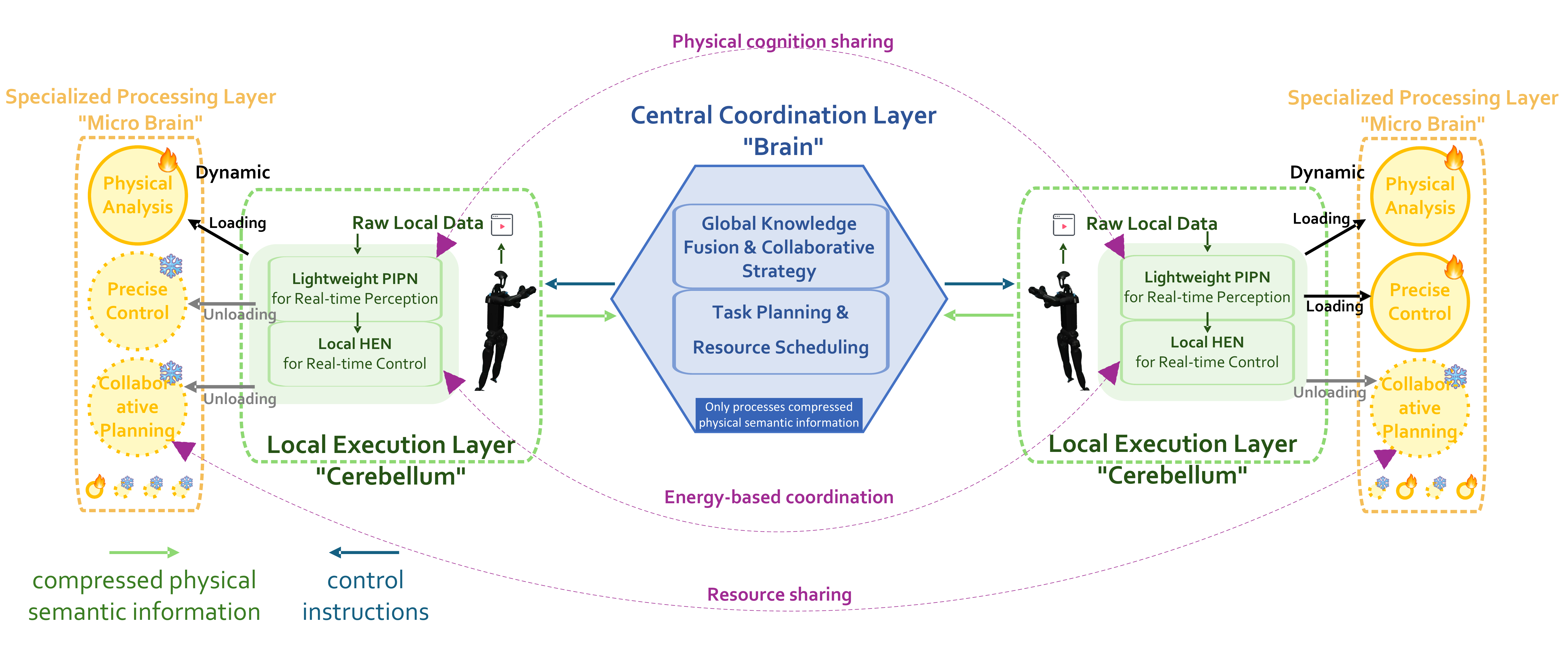}
		\caption{The distributed micro-brain architecture of PIPHEN, featuring three hierarchical levels. The central coordination layer ("Brain") handles global PIPN knowledge fusion and task planning, while only processing compressed physical-semantic information. The local execution layer ("Cerebellum") deploys lightweight PIPNs for edge intelligence and basic HENs for real-time control, processing local multimedia data on each robot. The specialized processing layer ("Micro-brain") provides dynamic capabilities through function-specific modules (e.g., physics analysis, precise control, collaborative planning) that can be dynamically loaded/unloaded and shared among robots based on task requirements. The green and blue arrows represent the information flow between layers, while the purple connections show resource sharing across robots.}
		\label{fig:micro_brain_architecture}
	\end{figure*}

	Our distributed micro-brain architecture, shown in Figure~\ref{fig:micro_brain_architecture}, features a three-level hierarchy. A central "Brain" performs global planning using compressed information, while local "Cerebellums" on each robot handle real-time control and edge intelligence. The architecture is enhanced by specialized "Micro-brain" modules that provide dynamic, shareable capabilities for complex tasks such as physics analysis and collaborative planning.

	\begin{table*}[htbp]
		\caption{Key Hyperparameters of the PIPHEN Framework}
		\label{tab:hyperparameters_appendix}
		\centering
		\begin{tabular}{lc}
			\toprule
			\textbf{Hyperparameter} & \textbf{Value} \\
			\midrule
			\multicolumn{2}{l}{\textbf{Training Parameters}} \\
			\quad Learning Rate & $1 \times 10^{-4}$ \\
			\quad Learning Rate Warmup & Cosine \\
			\quad Optimizer & AdamW \\
			\quad Weight Decay & $2 \times 10^{-4}$ \\
			\quad Batch Size & 128 \\
			\quad Discount Factor $\gamma$ & 0.96 \\
			\quad Polyak Update Rate $\tau$ & $8 \times 10^{-4}$ \\
			\quad Ensemble Size & 3 (for uncertainty estimation) \\
			\midrule
			\multicolumn{2}{l}{\textbf{Model Architecture}} \\
			\quad Hidden Layer Size & 768 \\
			\quad Physics Knowledge Graph Size & 512×512 \\
			\quad Hamiltonian Energy Network Layers & 4 \\
			\midrule
			\multicolumn{2}{l}{\textbf{Loss Function Weights}} \\
			\quad Physics Consistency Loss Weight $\lambda_{\text{phy}}$ & 0.1 \\
			\quad Energy Conservation Loss Weight $w_E$ & 1.0 \\
			\quad Momentum Conservation Loss Weight $w_M$ & 1.0 \\
			\quad HEN Energy Conservation Penalty Weight $\lambda$ & 0.05 \\
			\bottomrule
		\end{tabular}
	\end{table*}

	\section{Appendix Q: Sim-to-Real Transfer Methodology}
	\label{appendix:sim_to_real}
	As mentioned in the main text, our real-world deployment policy is trained entirely in simulation and transferred directly to the two XLeRobot manipulators with only minor tuning. This successful transfer is enabled by a two-stage approach combining System Identification (SysID) and Domain Randomization (DR).
	
	\subsection{System Identification (SysID)}
	The first stage involves creating a high-fidelity simulation baseline that closely matches real-world physics. We performed SysID on our XLeRobot setup to identify key physical parameters. This process included:
	\begin{itemize}
		\item \textbf{Robot Dynamics:} We identified the friction (viscous and Coulomb) and inertia parameters for each joint of the XLeRobot arms. This ensures that the torques required for movement in the simulation accurately reflect the real hardware.
		\item \textbf{Sensor Model:} We calibrated the camera intrinsics and characterized the noise profile (e.g., depth error characteristics) of the real RGB-D sensors, replicating this noise model in the simulation.
		\item \textbf{Object Properties:} We measured the precise mass, center of mass, and friction coefficients (static and dynamic) of the tableware (plates, cups, cutlery) used in the real-world task.
	\end{itemize}
	These identified parameters served as the "mean" values for our simulation environment.
	
	\subsection{Domain Randomization (DR)}
	The second stage involves introducing stochasticity during training to ensure the policy is robust to any errors in our system identification and to variations in the real world. Building upon the baseline SysID parameters, we randomized the following during simulation training:
	\begin{itemize}
		\item \textbf{Physics Parameters:} We applied randomization to the parameters identified in SysID. For example: object mass ($\pm$15\%), friction coefficients ($\pm$20\%), and robot joint friction ($\pm$10\%).
		\item \textbf{Visual Parameters:} To ensure robust perception (PIPN), we randomized lighting conditions (intensity, position, and color), shadow casting, and applied random textures to the table surface and robot grippers.
		\item \textbf{Control and Observation Delays:} We introduced random latencies (e.g., 5-25ms) to both the action commands and the sensor feedback to simulate real-world communication delays and processing time.
	\end{itemize}
	This combined SysID-DR approach creates a policy that is not overfitted to a specific simulation instance but is instead robust to a wide distribution of physical and visual parameters, enabling effective deployment in the complex, real-world environment.

\end{document}